\documentclass{article}
\usepackage{amsmath}
\usepackage{amssymb}
\usepackage{tabularx}
\usepackage{booktabs}
\usepackage{graphicx}
\usepackage{float}

\usepackage{titling}
\usepackage{hyperref}
\usepackage[section]{placeins}

\usepackage[style=apa, backend=biber]{biblatex}
\newcommand{\Hint}{\mathrm{H}_{\mathrm{int}}}
\newcommand{\dimeff}{\mathrm{dim}_{\mathrm{eff}}}
\newcommand{\Recov}{\mathrm{Recov}}

\providecommand{\keywords}[1]{\textbf{Keywords:} #1}
\pretitle{%
  \begin{center}
  \rule{\linewidth}{1pt}\\[0.8em]   
  \LARGE\bfseries                     
}
\posttitle{%
  \\[0.8em]\rule{\linewidth}{1pt}    
  \end{center}
}

\title{Intention Collapse: Intention-Level Metrics for Reasoning in Language Models}
\author{
  P.~M.~Vera \\
  School of Engineering and Applied Science \\
  George Washington University \\
  Washington, DC, USA \\
  \texttt{patricio.vera@gwu.edu}}
\date{} 

\begin{document}

\maketitle

\begin{abstract}
Language generation maps a rich, high-dimensional internal state to a single token sequence. We study this many-to-one mapping through the lens of \emph{intention collapse}: the projection from an internal intention space \(I\) to an external language space \(L\). We introduce three cheap, model-agnostic metrics computed on a pre-collapse state \(I\): (i) intention entropy \(H_{\mathrm{int}}(I)\), (ii) effective dimensionality \(\mathrm{dim}_{\mathrm{eff}}(I)\), and (iii) recoverability \(\mathrm{Recov}(I)\), operationalized as probe AUROC for predicting eventual success. We evaluate these metrics in a \(3\times 3\) study across models (Mistral-7B, LLaMA-3.1-8B, Qwen-2.5-7B) and benchmarks (GSM8K, ARC-Challenge, AQUA-RAT), comparing baseline, chain-of-thought (CoT), and a babble control (\(n=200\) items per cell). CoT increases average accuracy from \(34.2\%\) to \(47.3\%\) (+\(13.1\)pp), driven by large gains on GSM8K but consistent degradations on ARC-Challenge. Across models, CoT induces distinct entropy regimes relative to baseline, \(\Delta H = H_{\mathrm{int}}(\mathrm{CoT}) - H_{\mathrm{int}}(\mathrm{Base})\): Mistral shows \(\Delta H<0\) (lower-entropy CoT), whereas LLaMA shows \(\Delta H>0\) (higher-entropy CoT), highlighting heterogeneity in CoT-induced internal uncertainty. Finally, probe AUROC is significantly above chance in a subset of settings and can dissociate from behavioral accuracy (e.g., high AUROC alongside lower CoT accuracy on ARC-Challenge for Qwen), suggesting that informative internal signal is not always reliably converted into a final discrete decision under certain response formats.
\end{abstract}


\section{Introduction}
Every act of human communication begins in a space far richer than language itself. Before a speaker utters a single word, what exists is a high-dimensional, pre-linguistic manifold of memories, emotions, goals, analogies, and half-formed arguments. From this unbounded cloud of potential meanings, only one linear sequence of words is ultimately produced. The same internal state could have been expressed as ``I'm exhausted,'' ``I barely slept,'' ``My brain is mush today,'' or infinitely many other variants---yet exactly one utterance crystallizes at this particular moment.

We term this irreversible reduction \emph{intention collapse}: the process by which a cognitive system---biological or artificial---compresses a vast, implicit configuration of meaning into a single concrete linguistic message. Intention collapse is not a peripheral byproduct of language generation; it is a structural feature of any system that must commit a rich internal state to a discrete, serial surface.

Contemporary large language models (LLMs) exhibit a closely analogous pattern. Beneath the familiar formalism of next-token prediction lies a two-stage computation: first, the model integrates the input with parameters and context to form a high-dimensional internal representation; then, this representation is projected into a discrete token sequence. We formalize this as an intention state \(I\)---the prompt-conditioned latent representation immediately before the first token is emitted---and a collapse operator \(\kappa_\theta\) that maps \(I\) into language. The mapping is many-to-one and inherently lossy: information present in \(I\) may not survive the projection into the output.

This framing offers a unifying lens on a diverse landscape of reasoning-enhancing techniques. Chain-of-thought prompting \parencite{wei2022chain}, self-taught reasoning \parencite{zelikman2022star}, process reward models \parencite{lightman2023letsverify}, test-time training \parencite{sun2020ttt}, and self-refinement methods \parencite{madaan2023selfrefine} can be viewed as interventions that alter what the model ``brings to the brink'' of generation---i.e., the informational and geometric properties of \(I\) prior to collapse. Making \(I\) explicit lets us ask not only whether a method improves benchmark scores, but \emph{how} it reshapes the pre-collapse state and when such reshaping helps or harms performance.

To operationalize this framework, we propose three simple, model-agnostic intention metrics:
\begin{itemize}
  \item \textbf{Intention entropy \(H_{\text{int}}(I)\):} the Shannon entropy of the next-token distribution at the pre-collapse moment, measuring how concentrated or diffuse the model's immediate continuation is.
  \item \textbf{Effective dimensionality \(d_{\text{eff}}(I)\):} the participation ratio of PCA eigenvalues over pre-collapse hidden states, capturing the geometric richness of the intention representation.
  \item \textbf{Latent knowledge recoverability \(\text{Recov}(I; Z)\):} the performance of linear probes trained on \(I\) to predict task outcomes (e.g., answer correctness), quantifying information present pre-collapse that may fail to be expressed in the final output.
\end{itemize}
These metrics are cheap to compute, require no architectural modifications, and can be applied to any transformer-based model.

\paragraph{Empirical contributions.}
We instantiate the intention collapse framework in a systematic empirical study spanning:
\begin{itemize}
  \item \textbf{Three model families:} Mistral-7B-Instruct, Llama-3.1-8B-Instruct, and Qwen-2.5-7B-Instruct, covering diverse instruction-tuned model families.
  \item \textbf{Three reasoning benchmarks:} GSM8K (free-response grade-school math), ARC-Challenge (multiple-choice abstract reasoning), and AQUA-RAT (multiple-choice math word problems), chosen to contrast response formats and problem structures.
  \item \textbf{Three inference regimes:} direct-answer baseline, chain-of-thought (CoT), and a babble control for verbosity confounds.
\end{itemize}

Across this \(3\times3\) matrix, we find \emph{heterogeneous} effects that sharpen the intention-collapse perspective:
\begin{itemize}
  \item \textbf{CoT is not universally beneficial.} CoT yields large gains on GSM8K across all three models, but \emph{consistently degrades} performance on ARC-Challenge across all three models, and produces mixed outcomes on AQUA-RAT (including strong improvements for some models and substantial drops for others). This motivates separating changes in internal state from the reliability of the final discrete commitment.
  \item \textbf{CoT induces distinct internal entropy regimes across models.} When comparing CoT to baseline via \(\Delta H = H_{\text{int}}(\text{CoT}) - H_{\text{int}}(\text{Base})\), we observe qualitatively different regimes: Mistral operates under lower pre-collapse entropy in CoT (\(\Delta H<0\)), while LLaMA operates under higher pre-collapse entropy in CoT (\(\Delta H>0\)). These labels describe comparative uncertainty under CoT vs baseline and do not assume token-level temporal dynamics.
  \item \textbf{Response format is a major axis of variation.} Large within-domain gaps between free-response and multiple-choice settings (e.g., math in GSM8K vs math in AQUA-RAT) indicate strong response-format sensitivity, with cross-model reversals suggesting that format preference is not uniform across model families.
  \item \textbf{Recoverability signal is real but can dissociate from accuracy.} Linear probes trained on \(I\) achieve AUROC above chance in a subset of settings; strikingly, high AUROC can co-occur with degraded CoT accuracy in multiple-choice regimes, suggesting that an informative internal signal may fail to be reliably converted into a clean final decision under certain formats.
  \item \textbf{Babble helps isolate verbosity effects, but does not replicate CoT signatures.} Despite producing long outputs, babble does not systematically reproduce the same metric--performance relationships observed under CoT, supporting the view that CoT changes more than output length.
\end{itemize}

We also report important limitations. Item-level predictive power of \(H_{\text{int}}(I)\) is weak in several settings; probe performance varies across layers, tasks, and models; and the relationship between intention metrics and downstream accuracy is correlational rather than causal. Moreover, CoT substantially increases output length relative to baseline, introducing a compute confound that must be considered when interpreting accuracy gains.

\paragraph{Conceptual contributions.}
Beyond the empirical findings, we argue that intention collapse provides a useful vocabulary for reasoning about LLM behavior:
\begin{itemize}
  \item It reframes diverse inference-time methods as interventions on a common object (\(I\)), enabling direct comparison via shared metrics.
  \item It connects to parallel literatures on latent knowledge elicitation \parencite{burns2022discovering}, pre-generation adequacy \parencite{stengel2024clotho}, and uncertainty estimation \parencite{kuhn2023semantic}, offering a conceptual bridge across these threads.
  \item It suggests concrete research directions, including state-dependent decoding policies that adapt sampling temperature based on \(I\), and early-exit strategies that use \(\text{Recov}(I; Z)\) to avoid unnecessary computation.
\end{itemize}

We do not claim that LLMs ``have intentions'' in the conscious or phenomenological sense. Rather, we argue that both biological and artificial systems exhibit a shared computational shape: rich internal dynamics followed by a sharply reduced verbal surface. Intention collapse is meant to capture this structural analogy without erasing the many substantive differences between human cognition and neural network computation. While the notion of intention collapse is modality-agnostic---applying equally to text, images, or multimodal inputs---we focus here on text-only benchmarks and leave multimodal validation to future work.

\paragraph{Outline.}
Section~2 operationalizes intention collapse with precise definitions and extraction protocols. Section~3 situates our framework within contemporary reasoning techniques. Section~4 presents the experimental study. Section~5 discusses related work. Section~6 addresses limitations, and Section~7 concludes with directions for future research.

\newpage

\section{Operationalizing Intention Collapse}
To move intention collapse from metaphor to measurement, we require (i) a precise definition of the pre-collapse intention state \(I\) in transformer LLMs, (ii) a reproducible extraction point, and (iii) metrics that quantify uncertainty, geometry, and recoverable task-relevant signal in \(I\).
We instantiate these definitions in a controlled \(3\times 3\) empirical design across models (Mistral-7B-Instruct, Llama-3.1-8B-Instruct, Qwen-2.5-7B-Instruct) and benchmarks (GSM8K, ARC-Challenge, AQUA-RAT), under three inference regimes (Baseline, CoT, Babble), using \(n=200\) items per model--benchmark cell. Throughout, we treat \(I\) as a \emph{pre-generation} object: it is extracted after prompt conditioning and before the first output token is selected.

\subsection{Intention State: Definition and Extraction Protocol}

\textbf{Definition 1 (Intention State).}
For a transformer-based language model with \(L\) layers processing a prompt \(x_{1:S}\), we define the \emph{intention state} \(I\) as the collection of hidden representations at the first decoding step, immediately before the first output token is selected:
\begin{equation}
I \triangleq \{h^{(\ell)}_{\text{pre}}\}_{\ell \in \mathcal{L}} \in \mathbb{R}^{|\mathcal{L}| \times d},
\end{equation}
where \(h^{(\ell)}_{\text{pre}} \in \mathbb{R}^{d}\) is the hidden state at layer \(\ell\) at the final prompt position (the position from which the first output token will be generated), and \(\mathcal{L} \subseteq \{1,\ldots,L\}\) is a selected subset of layers.

\paragraph{Extraction protocol.}
We extract \(I\) at a fixed, well-defined moment:
\begin{enumerate}
  \item The model processes the complete prompt \(x_{1:S}\) (including any condition-specific instruction, e.g., Baseline vs CoT vs Babble).
  \item A forward pass computes hidden states across all layers.
  \item We record \(h^{(\ell)}_{\text{pre}}\) at the last prompt position for each \(\ell \in \mathcal{L}\).
\end{enumerate}
\emph{Collapse} begins when the model converts this state into a discrete choice for the first output token \(y_1\) (via greedy selection or sampling).

This definition ensures that \(I\) is strictly \emph{pre-collapse}: it captures the model's internal state after prompt conditioning but before any discrete token commitment. The same extraction point is used for all experimental regimes, enabling controlled comparison of how prompt strategy shapes the state right before generation begins.

\paragraph{Clarification on internal computation and ``thinking time''.}
Even when a model emits its first token immediately, it still performs a nontrivial internal computation from layer \(1\) to layer \(L\) on the prompt. In our framework, this layer-wise computation constitutes the pre-collapse processing that culminates in \(I\). Prompting strategies such as CoT do \emph{not} require hidden ``thinking tokens'' to influence \(I\): they can reshape \(I\) purely through prompt conditioning, which changes the internal representation before the first token is selected. (Any additional reasoning tokens emitted after \(y_1\) are, by definition, post-collapse and are analyzed separately when relevant.)

\paragraph{Modality-agnostic formulation.}
While we evaluate the framework on text-only tasks, the definition is modality-agnostic: in multimodal models, \(I\) would be extracted after joint encoding of the available inputs but still immediately before the first generated token. We leave cross-modality validation to future work.

\subsection{Intention Metrics}
We propose three complementary metrics that quantify different aspects of the pre-collapse intention state. All are inexpensive, require no architectural changes, and can be computed from standard forward-pass outputs.

\subsubsection{Intention Entropy \texorpdfstring{$H_{\text{int}}(I)$}{Hint(I)}}
\textbf{Definition 2 (Intention Entropy).}
Intention entropy is the Shannon entropy of the next-token distribution at the pre-collapse moment:
\begin{equation}
H_{\text{int}}(I) \triangleq H\bigl(p_\theta(y_1 \mid x_{1:S})\bigr)
= -\sum_{v \in \mathcal{V}} p_\theta(y_1 = v \mid x_{1:S}) \log_2 p_\theta(y_1 = v \mid x_{1:S}),
\end{equation}
where \(\mathcal{V}\) is the vocabulary. Lower entropy corresponds to a more concentrated continuation distribution at the moment of commitment.

\paragraph{Computation.}
We compute \(H_{\text{int}}(I)\) from the logits at the last prompt position, immediately before selecting \(y_1\). Entropy is reported in bits (base 2) using numerically stable log-sum-exp computations.

\paragraph{Scope and interpretation.}
Throughout, \(H_{\text{int}}(I)\) is an \emph{unconstrained} next-token entropy over the full vocabulary. It is therefore best interpreted as a measure of \emph{local generative dispersion} at the pre-collapse boundary, not as a calibrated measure of uncertainty over task-specific answer choices. In particular, \(H_{\text{int}}(I)\) can be sensitive to the lexical and tokenization details at the end of the prompt (e.g., small changes in suffixes or separators). Accordingly, we use changes in \(H_{\text{int}}\) primarily to characterize whether a model's pre-collapse distribution becomes more concentrated or more diffuse across conditions, rather than to directly quantify decision uncertainty.

\paragraph{Decision-space--aligned alternatives (MCQ).}
For multiple-choice benchmarks, a more decision-aligned diagnostic would restrict the distribution to option tokens (e.g., \(\{A,B,C,D,E\}\)) and report an \emph{option-normalized entropy} and pre-collapse \emph{logit margins} between the top options. These quantities require logging per-option log-probabilities at the same pre-first-token boundary. Our current checkpoint format stores aggregated entropy values but not per-option log-probabilities, so we do not report option-normalized entropy or margins in this revision; we instead scope claims about \(H_{\text{int}}\) to the unconstrained distribution and include this logging as future work (Sec.~\ref{sec:limitations}).

\subsubsection{Effective Dimensionality \texorpdfstring{$d_{\text{eff}}(I)$}{deff(I)}}
\textbf{Definition 3 (Per-layer Effective Dimensionality).}
For a collection of \(N\) intention states \(\{I_i\}_{i=1}^N\) across examples, the per-layer effective dimensionality at layer \(\ell\) is the \emph{participation ratio} of the PCA eigenvalue spectrum:
\begin{equation}
d_{\text{eff}}^{(\ell)} \triangleq
\frac{\left(\sum_{j=1}^{d} \lambda_{j}^{(\ell)}\right)^2}{
\sum_{j=1}^{d} \left(\lambda_{j}^{(\ell)}\right)^2},
\end{equation}
where \(\{\lambda_{j}^{(\ell)}\}_{j=1}^{d}\) are the eigenvalues of the sample covariance matrix of hidden states at layer \(\ell\).

\paragraph{PCA protocol.}
For each layer \(\ell \in \mathcal{L}\):
\begin{enumerate}
  \item Collect hidden states \(\{h^{(\ell)}_{\text{pre},i}\}_{i=1}^{N}\) across \(N\) examples.
  \item Construct the data matrix \(X^{(\ell)} \in \mathbb{R}^{N \times d}\) with each row being one example's hidden state.
  \item Center by subtracting the column mean:
  \begin{equation}
    \tilde{X}^{(\ell)} = X^{(\ell)} - \mathbf{1}\mu^{(\ell)\top},
    \qquad
    \mu^{(\ell)} = \frac{1}{N}\sum_{i=1}^{N} h^{(\ell)}_{\text{pre},i}.
  \end{equation}
  \item Compute the sample covariance and extract eigenvalues:
  \begin{equation}
    C^{(\ell)} = \frac{1}{N-1}\tilde{X}^{(\ell)\top}\tilde{X}^{(\ell)}.
  \end{equation}
  \item Apply the participation ratio formula above.
\end{enumerate}

\paragraph{Finite-sample ceiling and stability.}
Because \(C^{(\ell)}\) is estimated from \(N\) samples in \(d \gg N\) dimensions, its rank is bounded by \(\mathrm{rank}(C^{(\ell)}) \le N-1\), which imposes a finite-sample ceiling on any spectrum-based dimensionality estimate. To reduce sensitivity to this regime, we (i) compute \(d_{\text{eff}}^{(\ell)}\) \emph{per layer} rather than via concatenation-based ``global PCA'' across heterogeneous layer spaces, and (ii) report subsampling stability of \(d_{\text{eff}}\) across \(N \in \{25,50,100,150,200\}\) (Appendix~\ref{app:deff_robustness}). The qualitative cross-condition patterns emphasized in the paper are stable under these subsamples.

\textbf{Definition 4 (Layer-aggregated Effective Dimensionality).}
To aggregate across layers, we compute a variance-weighted average:
\begin{equation}
d_{\text{eff}}^{\text{agg}} \triangleq \sum_{\ell \in \mathcal{L}} w_{\ell}\, d_{\text{eff}}^{(\ell)},
\qquad
w_{\ell} = \frac{\mathrm{tr}(C^{(\ell)})}{\sum_{\ell' \in \mathcal{L}} \mathrm{tr}(C^{(\ell')})}.
\end{equation}
This avoids concatenation-based ``global PCA'' across heterogeneous layer spaces and yields a stable summary index.

\paragraph{Cross-checks with intrinsic-dimension estimators.}
To further validate that observed shifts are not artifacts of participation-ratio estimation in the \(d \gg N\) regime, we additionally compare trends against intrinsic-dimension estimators based on nearest-neighbor statistics (TwoNN) and maximum-likelihood methods (Levina--Bickel), reported in Appendix~\ref{app:deff_robustness}. These estimators operate on local distance structure and provide complementary evidence to spectrum-based \(d_{\text{eff}}\).

\paragraph{Interpretation.}
Higher \(d_{\text{eff}}\) indicates that the intention representation spans a higher-dimensional subspace, consistent with a less constrained pre-collapse state under the model's internal geometry at the extraction boundary.

\subsubsection{Latent Knowledge Recoverability \texorpdfstring{$\text{Recov}(I; Z)$}{Recov(I; Z)}}
\textbf{Definition 5 (Recoverability).}
For a target variable \(Z\) (e.g., answer correctness), recoverability from intention state \(I\) is the predictive performance of a linear probe trained on \(I\):
\begin{equation}
\text{Recov}(I; Z) \triangleq \text{AUROC}\bigl(f_{\phi}(I), Z\bigr),
\end{equation}
where \(f_{\phi}(I)=\sigma\!\left(W \cdot \mathrm{vec}(I) + b\right)\) is an \(\ell_2\)-regularized logistic regression classifier and \(\sigma(\cdot)\) is the sigmoid.

\paragraph{Probe training protocol.}
To reduce leakage and preserve item-level independence:
\begin{itemize}
  \item \textbf{Split by item (not by token).} Each problem is assigned to exactly one of train/validation/test (60\%/20\%/20\%). When multiple inference regimes are evaluated for the same item, all regimes remain in the same split.
  \item \textbf{Feature construction.} For each example, concatenate hidden states across selected layers:
  \begin{equation}
    \mathrm{vec}(I) = \bigl[h^{(\ell_1)}_{\text{pre}}; h^{(\ell_2)}_{\text{pre}}; \ldots\bigr] \in \mathbb{R}^{|\mathcal{L}| \cdot d}.
  \end{equation}
  \item \textbf{Normalization.} Apply z-score normalization using training-set statistics only:
  \begin{equation}
    \tilde{x}_j = \frac{x_j - \mu^{\text{train}}_j}{\sigma^{\text{train}}_j}.
  \end{equation}
  \item \textbf{Regularization.} Select \(C \in \{10^{-4},10^{-3},\ldots,10^{2}\}\) by validation AUROC; report test AUROC for the selected setting.
\end{itemize}

\paragraph{Evaluation and uncertainty.}
We report AUROC as the primary recoverability metric (robust to class imbalance), along with 95\% confidence intervals computed via bootstrap resampling (1000 resamples). In addition to test AUROC, we also report train vs.\ test AUROC (Appendix~\ref{app:probe_robustness}) to quantify potential overfitting under the high feature-to-sample ratio (\(n=200\) per cell). We interpret probe estimates conservatively and emphasize cross-condition and cross-format comparisons.

\paragraph{Sanity-check baselines.}
To validate that recoverability is not an artifact of spurious correlations or feature scaling, we include two sanity checks (Appendix~\ref{app:probe_robustness}): (i) a constant-score baseline (which yields AUROC \(=0.5\)), and (ii) \textbf{label-shuffle probes}, where training labels are randomly permuted and the probe is retrained; AUROC should collapse to chance if the pipeline is well-specified.

\paragraph{Cross-regime transfer.}
To test whether recoverable information corresponds to shared predictive axes or regime-specific encodings, we additionally evaluate \textbf{cross-regime transfer}: train probes on intention states from one regime (e.g., Baseline) and test on another (e.g., CoT), and vice versa, while preserving the same item splits. We summarize these results as transfer matrices (Appendix~\ref{app:probe_transfer}). Systematic degradation under transfer indicates representational shift across regimes even when within-regime recoverability remains above chance.

\paragraph{Degrees of freedom and layer selection.}
Layer selection is treated as a validation-only hyperparameter. To limit researcher degrees of freedom, we also report results for a fixed pre-specified layer set used across all conditions and models (Appendix~\ref{app:probe_robustness}). Where applicable, we compare probes trained on pre-collapse intention states against probes trained on post-collapse representations.

\paragraph{Probe robustness (summary).}
Given the high feature-to-sample ratio (tens of thousands of features for \(n=200\) items), we evaluate probe robustness via train--test AUROC gaps and a label-shuffle sanity check. Shuffled-label AUROC remains near chance across cells, supporting pipeline validity, while several cells exhibit large train--test gaps, motivating conservative interpretation of probe results. Full robustness results are reported in Appendix~\ref{app:probe_robustness}.

\paragraph{Cross-regime probe transfer indicates representational shift.}
We additionally evaluate cross-regime probe transfer (train on Baseline and test on CoT/Babble, and vice versa) while preserving item splits. Transfer AUROC often degrades relative to within-regime AUROC (Appendix~\ref{app:probe_transfer}), indicating that linearly accessible signals are frequently encoded along regime-specific axes. This supports the intention-collapse view that prompting regimes can shift the geometry of \(I\) even when within-regime recoverability remains above chance.

\subsection{Experimental Conditions}
We evaluate intention metrics under three inference regimes, holding decoding parameters constant (greedy decoding, temperature \(T=0\)) to isolate the effect of prompting strategy:

\begin{table}[t]
\centering
\begin{tabular}{p{0.18\linewidth} p{0.50\linewidth} p{0.28\linewidth}}
\hline
\textbf{Condition} & \textbf{Prompt Strategy} & \textbf{Purpose} \\
\hline
Baseline & Direct answer instruction & Minimal reasoning; immediate commitment \\
CoT (Enhanced) & ``Think step by step'' instruction & Elicit explicit intermediate reasoning \\
Babble & Stream-of-consciousness instruction (unrelated to solving) & Length-oriented control for verbosity confounds \\
\hline
\end{tabular}
\caption{Inference regimes used to evaluate intention metrics.}
\label{tab:conditions}
\end{table}

\paragraph{Critical note on extraction timing.}
In all three conditions, we extract \(I\) at the \emph{same} computational moment: after prompt processing and before the first output token. The regimes differ in the instructions embedded in the prompt---and thus can shape \(I\) through conditioning---but the extraction point is identical. This design cleanly separates (i) differences attributable to pre-collapse conditioning from (ii) post-collapse dynamics during generation.

\paragraph{Connection to empirical signatures.}
In the \(3\times 3\) study reported in Section~4, this protocol reveals that (a) CoT effects on accuracy are strongly task- and format-dependent (large gains on GSM8K, consistent degradations on ARC-Challenge, mixed outcomes on AQUA-RAT), (b) CoT induces different entropy \emph{regimes} across model families when compared to baseline via \(\Delta H\), and (c) recoverability (probe AUROC) can remain above chance even when CoT accuracy declines in multiple-choice settings---highlighting a potential dissociation between internal separability and final discrete commitment. These findings motivate treating \(I\) as an object of study in its own right.


\section{Relationship to Contemporary Reasoning Techniques}
\label{sec:relationship_reasoning}

A growing body of work between roughly 2022 and 2025 has aimed to improve LLM reasoning by altering what the model brings to the moment of commitment, rather than only post-hoc reshaping the final answer. These approaches differ in mechanism---prompting, supervision, test-time adaptation, or search---but share a common structure: they modify the information content, geometry, or uncertainty of the model's internal state prior to emitting an answer. In our framework, they are naturally reinterpreted as \textbf{interventions on the intention state \(I\)} and on how collapse maps \(I\) to a discrete output.

This reframing is not merely taxonomic. In our \(3\times 3\) study (three model families; three benchmarks spanning free-response and multiple-choice formats), we find that interventions commonly grouped under ``reasoning'' can yield sharply different behavioral outcomes depending on response format and model family. This motivates treating \(I\) as an empirical object: if two methods both increase output length or surface ``reasoning'', they need not induce the same pre-collapse regime, nor the same reliability of final discrete commitment.

\begin{table}[t]
    \centering
    \small
    \begin{tabularx}{\textwidth}{@{} >{\raggedright\arraybackslash}p{0.22\textwidth} X >{\raggedright\arraybackslash}p{0.22\textwidth} @{}}
        \toprule
        \textbf{Method family} & \textbf{Intervention on \(I\) (intention-collapse lens)} & \textbf{Key reference} \\
        \midrule

        Self-generated rationales (STaR) &
        Re-shapes the training distribution so that pre-collapse intention states \(I\) more often encode multi-step, solution-bearing structure before any discrete commitment. &
        \parencite{zelikman2022star} \\
        \addlinespace 

        Internal thought channels (Quiet-STaR) &
        Adds latent token-level “thought” computations that modify representations prior to emitting the next visible token, effectively enlarging the internal pre-commitment workspace. &
        \parencite{zelikman2024quietstar} \\
        \addlinespace

        Process supervision (PRMs) &
        Imposes step-level constraints that bias trajectories through intention space toward intermediate states judged valid, thereby narrowing collapse toward higher-quality regions of \(I\). &
        \parencite{lightman2023letsverify, ma2023stepreward, setlur2024rewarding} \\
        \addlinespace

        Test-time training (TTT) &
        Makes intention formation instance-adaptive by updating a subset of parameters at inference time, producing \(I(x;\theta')\) tailored to the current input before generation. &
        \parencite{gandelsman2022tttmae, akyurek2025tttabstract, mitttt2024overview} \\
        \addlinespace

        Iterative refinement / search (Self-Refine, MCTS, best-of-\(N\)) &
        Separates exploration from commitment by generating multiple candidate intention states (and/or candidate outputs) and selecting a final collapse using an internal or external scoring signal. &
        \parencite{madaan2023selfrefine} \\

        \bottomrule
    \end{tabularx}
    \caption{Representative reasoning techniques reinterpreted as interventions on the pre-collapse intention state \(I\). The shared mechanism is not surface verbosity, but structured modification of (i) the formation of \(I\) and/or (ii) the commitment policy mapping \(I\) into a discrete output.}
    \label{tab:reasoning_methods}
\end{table}

In what follows, we briefly survey several representative approaches and reformulate them as ways of enriching, refining, or constraining \(I\) before the irreversible projection into a concrete utterance. Our goal is not an exhaustive taxonomy, but a unifying language that makes testable predictions about how different techniques should affect \(H_{\text{int}}(I)\), \(d_{\text{eff}}(I)\), and \(\text{Recov}(I;Z)\).

\subsection{Self-generated rationales: STaR and Quiet-STaR}

\paragraph{STaR.}
Self-Taught Reasoner (STaR) \parencite{zelikman2022star} trains language models to generate chain-of-thought rationales and then fine-tunes them on those rationales that lead to correct answers. From our perspective, STaR modifies the training distribution so that the model recurrently visits richer multi-step computation states before collapse. Rather than mapping problems directly to short answers, the model learns a mapping
\[
x \;\mapsto\; I_{\text{rationale}}(x) \;\mapsto\; y,
\]
where \(I_{\text{rationale}}(x)\) denotes an intention representation shaped by structured intermediate reasoning (partially externalized as a rationale).

In the language of Section~2, STaR should alter both geometry and information: it can increase \(d_{\text{eff}}(I)\) by expanding the diversity of reachable states, and increase \(\text{Recov}(I;Z)\) by concentrating probability mass on intention regions predictive of successful collapse. Crucially, this does not imply that longer rationales always help: as our empirical results later show, reasoning-like prompting can degrade accuracy in multiple-choice formats even when internal signal remains recoverable, underscoring the distinction between internal separability and final commitment.

\paragraph{Quiet-STaR.}
Quiet-STaR \parencite{zelikman2024quietstar} extends this idea by training models to produce \emph{token-level internal thoughts} that are never shown to the user but improve future-token prediction. Rather than emitting a visible chain-of-thought, the model generates an internal monologue that shapes hidden states before producing the next token. This is close to a literal instantiation of an explicit intention phase:
\[
x, \theta \;\longrightarrow\; I_{1}, I_{2}, \dots, I_{T_{\text{think}}}
\;\longrightarrow\; \kappa(I_{T_{\text{think}}}, \xi) = y.
\]
In our framework, such mechanisms should increase \(\text{Recov}(I;Z)\) for downstream outcomes by making relevant latent variables more linearly accessible prior to commitment, while potentially changing entropy regimes by stabilizing or diversifying the continuation distribution depending on task demands.

\subsection{Process supervision: Process Reward Models}

Process Reward Models (PRMs) and related work on process supervision
\parencite{lightman2023letsverify, ma2023stepreward, setlur2024rewarding}
provide feedback not only on the final answer but on the structure of the reasoning process itself. Instead of rewarding only correctness of \(y\), these methods assign rewards to intermediate steps, partial derivations, or proof states.

Viewed through our lens, PRMs define constraints over trajectories in intention space: some internal paths are marked as valid and others as invalid, and learning updates \(\theta\) to increase the probability of trajectories that pass through high-reward regions of \(I\). This suggests testable signatures in our metrics. For example, one might observe increased recoverability of correctness from \(I\) (higher probe AUROC) even when surface behavior varies by response format; conversely, one might observe that tighter process constraints reduce \(H_{\text{int}}(I)\) on tasks where collapse should be decisive, but increase \(d_{\text{eff}}(I)\) when multiple solution strategies remain viable.

\subsection{Test-time training and slow adaptation of \(\theta\)}

Test-Time Training (TTT) and related methods perform dynamic adaptation of model parameters (or adapters) during inference, often on unlabeled data from the test distribution
\parencite{gandelsman2022tttmae, akyurek2025tttabstract, mitttt2024overview}.
Applied to language and abstract reasoning, updating a small subset of parameters at test-time can improve performance under distribution shift.

In our framework, TTT makes parts of \(\theta\)---and hence parts of intention formation---instance-specific. The intention state \(I(x;\theta)\) is no longer produced solely by a fixed parameterization but by a slowly adapted one:
\[
\theta \;\xrightarrow{\;\text{TTT on } x\;}\; \theta'
\;\xrightarrow{\;\text{forward}\;}\; I(x; \theta') \;\xrightarrow{\;\kappa\;}\; y.
\]
This offers a clean metric-level prediction: compared to baseline prompting, TTT may change not only output accuracy but the recoverability and entropy of \(I\), potentially reducing uncertainty on ambiguous inputs or increasing the accessibility of task-relevant latent variables.

\subsection{Iterative refinement: Self-Refine, MCTS and best-of-\(N\)}

Another family of techniques performs iterative refinement of candidate solutions, often using the model to critique and revise its own outputs. Self-Refine \parencite{madaan2023selfrefine} prompts the model to first produce an initial draft and then generate feedback and an improved version, repeating as needed. Monte Carlo Tree Search (MCTS) over reasoning trajectories, best-of-\(N\) sampling, and related search-based methods similarly expand a set of candidates and select the best one according to a scoring signal.

In our terms, these methods explicitly decouple \emph{exploration in intention space} from \emph{final commitment}. The model traverses multiple candidate intention trajectories
\[
I^{(1)}, I^{(2)}, \dots, I^{(N)},
\]
each giving rise to a candidate output \(y^{(i)} = \kappa(I^{(i)}, \xi^{(i)})\), and then selects a winner using either the model itself or an external scorer. Two implications follow for our metrics:
\begin{itemize}
  \item The effective dimensionality of the ensemble \(\{I^{(i)}\}\) can exceed that of a single forward pass, especially early in the search.
  \item The final commitment can be made effectively low-entropy even if exploration uses high-variance sampling, because selection collapses variability onto a single chosen candidate.
\end{itemize}
This decoupling highlights why output length alone is an insufficient proxy for reasoning: methods with similar verbosity can induce different pre-collapse regimes, and conversely, strong internal signal (high \(\text{Recov}(I;Z)\)) can fail to translate into correct discrete choices under certain response formats, as our multiple-choice results later illustrate.


\section{Experimental Study and Research Agenda}
\label{sec:experiments}

While intention collapse is a conceptual framing, its usefulness hinges on whether pre-collapse measurements reveal stable, interpretable structure across models, tasks, and inference regimes. In this section we report a controlled \(3\times3\) empirical study instantiating the core protocol, and then outline extensions that remain as forward-looking components of the broader agenda. Our emphasis is on (i) how prompting strategies reshape the pre-collapse intention state \(I\), (ii) when such reshaping improves downstream accuracy, and (iii) whether \(I\) contains recoverable task-relevant signal that is not reliably expressed in the final output.

We expected signatures based on patterns observed in related literature, such as memorization vs.\ genuine reasoning in math benchmarks \parencite{cobbe2021gsm8k, hendrycks2021math}
and elicitation of hidden knowledge \parencite{burns2023quirky}. All protocols can be run on accessible models like Llama-3-8B, Qwen-2-7B, and
Mistral-7B variants \parencite{meta2024llama, qwen2024, mistral2023}, using
libraries such as Hugging Face Transformers for activation extraction and
scikit-learn for PCA and probing. Benchmarks include GSM8K (grade-school
math) \parencite{cobbe2021gsm8k}, MATH (competition-level math)
\parencite{hendrycks2021math}, and ARC (abstraction and reasoning)
\parencite{chollet2019arc}, which test multi-step reasoning and have seen recent
scrutiny for distinguishing rote memorization from flexible cognition
\parencite{deepeval2025gsm8k, medium2025reasoning}.
(Our primary reported matrix uses GSM8K, ARC-Challenge, and AQUA-RAT; MATH remains a natural extension.)

\subsection{Core Protocol: \(3\times3\) Matrix Across Models and Benchmarks}
\label{sec:core_protocol}

\paragraph{Design.}
We evaluate three instruction-tuned open-weight models---Mistral-7B-Instruct, Llama-3.1-8B-Instruct, and Qwen-2.5-7B-Instruct---on three benchmarks spanning free-response and multiple-choice formats: GSM8K (free-response math), ARC-Challenge (multiple-choice abstract reasoning), and AQUA-RAT (multiple-choice math word problems). For each model--benchmark cell, we sample \(n=200\) items and run three inference regimes: Baseline (direct answer), CoT (``think step by step''), and Babble (length-oriented control). We hold decoding fixed (greedy, \(T=0\)) to isolate prompt-conditioning effects.

\paragraph{Why these benchmarks.}
GSM8K \parencite{cobbe2021gsm8k} provides a free-response arithmetic reasoning setting with explicit final answers. ARC-Challenge \parencite{chollet2019arc} stresses discrete-choice abstract reasoning under a multiple-choice interface. AQUA-RAT provides a multiple-choice math setting, enabling within-domain comparison of response format (free-response vs.\ multiple-choice) while keeping problem content broadly mathematical.

\paragraph{Prompts and answer extraction.}
We use a minimal prompt template per regime: Baseline requests a direct final answer; CoT requests step-by-step reasoning followed by a final answer; Babble requests long continuation unrelated to solving. For multiple-choice benchmarks, we extract the final option letter from the model output using a conservative parser (single-choice extraction). For free-response (GSM8K), we extract the final numeric answer using standard GSM8K-style normalization. We treat extraction failures as incorrect, and we record compliance diagnostics (e.g., whether a valid option token was produced) for auditing format effects.

\paragraph{Pre-collapse extraction and metrics.}
For each run we extract the intention state \(I\) exactly as defined in Section~2: hidden representations at the last prompt position, immediately before selecting the first output token. We compute (i) intention entropy \(H_{\text{int}}(I)\), (ii) effective dimensionality \(d_{\text{eff}}(I)\) via per-layer PCA participation ratios with variance-weighted aggregation, and (iii) recoverability \(\text{Recov}(I;Z)\) using linear probes predicting eventual correctness \(Z\) from \(\mathrm{vec}(I)\).

\paragraph{Probing protocol and uncertainty estimation.}
For each model--benchmark cell and regime, we train \(\ell_2\)-regularized logistic regression probes with item-level splits (60/20/20) and report AUROC with bootstrap 95\% confidence intervals (1000 resamples). Given \(n=200\) per cell, we interpret probe results conservatively and focus on robust cross-condition patterns rather than overfitting to individual cells.

\subsection{Main Results: Accuracy, Entropy Regimes, and Recoverability}
\label{sec:main_results}

\paragraph{Accuracy is strongly format-dependent; CoT is not universally beneficial.}
Across the \(3\times3\) matrix, CoT increases average accuracy from \(34.2\%\) (Baseline) to \(47.3\%\) (+\(13.1\)pp), driven by large gains on GSM8K for all three models. However, CoT \emph{consistently degrades} performance on ARC-Challenge for all three models, and yields mixed outcomes on AQUA-RAT (including strong improvements for some models and substantial drops for others). These results motivate a central distinction in the intention-collapse view: changing the pre-collapse state and producing longer ``reasoning'' traces does not guarantee a reliable final discrete commitment under all response formats.

\begin{figure}[t]
  \centering
  \includegraphics[width=0.92\linewidth]{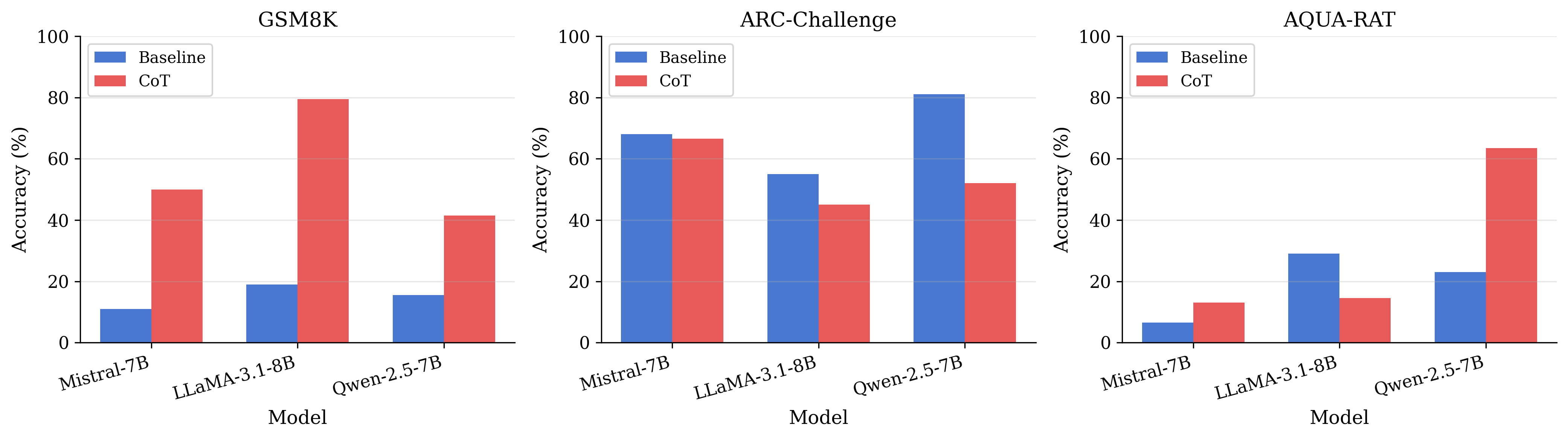}
  \caption{Accuracy across Baseline, CoT, and Babble for the \(3\times3\) model--benchmark matrix (\(n=200\) per cell). CoT yields large gains on GSM8K across models, but degrades ARC-Challenge across all three models, illustrating strong response-format dependence.}
  \label{fig:accuracy_comparison}
\end{figure}

\paragraph{CoT induces distinct entropy regimes across model families.}
We quantify regime shifts via \(\Delta H = H_{\text{int}}(\text{CoT}) - H_{\text{int}}(\text{Base})\). We observe qualitatively different entropy regimes across model families: Mistral operates under lower pre-collapse entropy in CoT (\(\Delta H<0\)), whereas LLaMA operates under higher pre-collapse entropy in CoT (\(\Delta H>0\)). (For Qwen, \(\Delta H\) is available for a subset of benchmarks due to missing baseline entropy logs in some runs; we report available cells and treat missingness explicitly as a limitation.) Importantly, these regimes describe comparative uncertainty under CoT vs.\ baseline and do not imply token-level temporal dynamics.

\begin{figure}[t]
  \centering
  \includegraphics[width=0.92\linewidth]{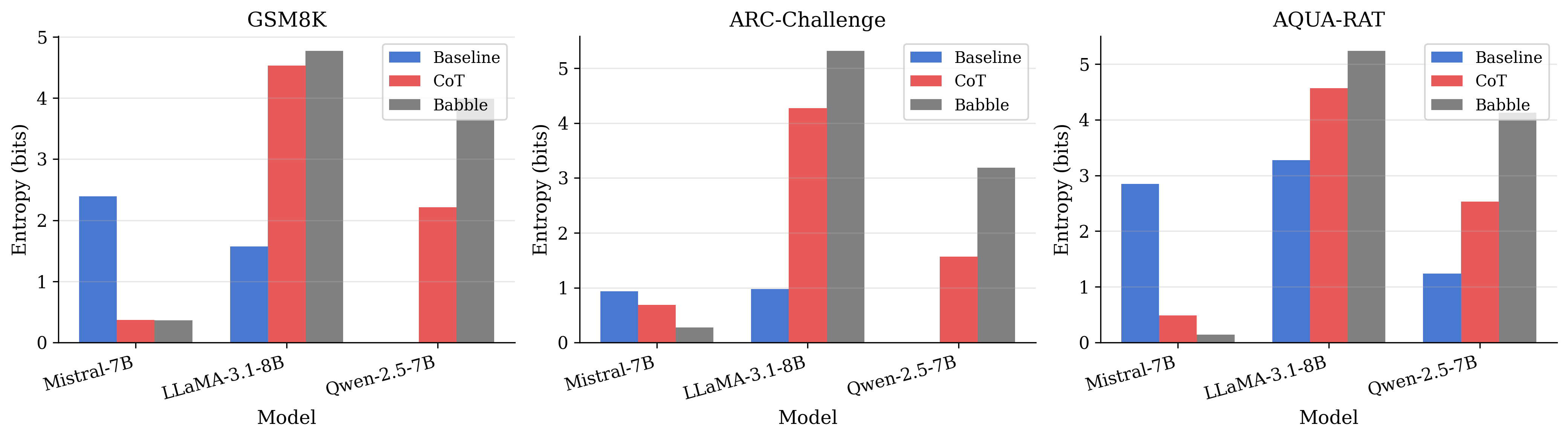}
  \caption{Pre-collapse intention entropy \(H_{\text{int}}(I)\) by condition (Baseline, CoT, Babble) across models and benchmarks.}
  \label{fig:entropy_by_condition}
\end{figure}

\begin{figure}[t]
  \centering
  \includegraphics[width=0.92\linewidth]{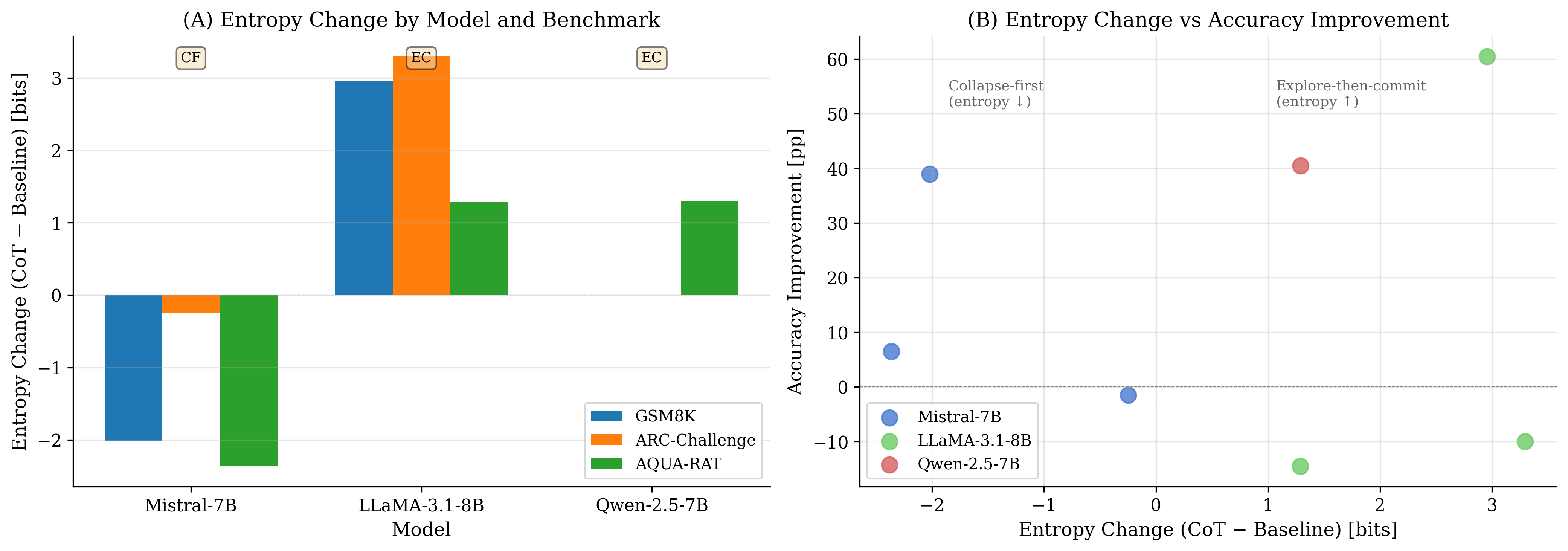}
  \caption{Entropy regime shifts under CoT, \(\Delta H = H_{\text{int}}(\text{CoT})-H_{\text{int}}(\text{Base})\), highlighting model-family heterogeneity (e.g., Mistral \(\Delta H<0\) vs.\ LLaMA \(\Delta H>0\)).}
  \label{fig:entropy_patterns}
\end{figure}

\paragraph{Within math-like tasks, response format is a major axis of variation.}
A salient pattern emerges when comparing math-like settings under different response formats: GSM8K (free-response) versus AQUA-RAT (multiple-choice). We observe large within-domain gaps and cross-model reversals consistent with strong response-format sensitivity. This motivates treating multiple-choice failure modes as potentially distinct from domain competence, and treating answer compliance (valid option selection) as a first-class diagnostic.

\begin{figure}[t]
  \centering
  \includegraphics[width=0.92\linewidth]{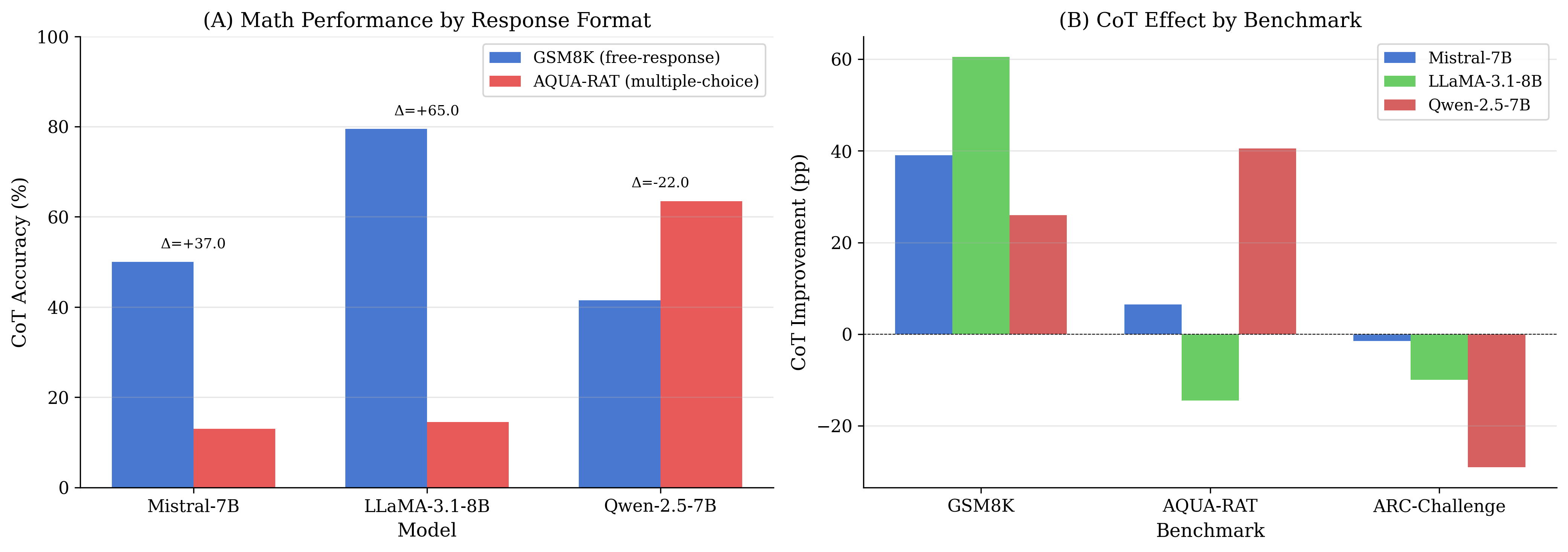}
  \caption{Format sensitivity within math-like tasks: free-response (GSM8K) vs.\ multiple-choice (AQUA-RAT). Cross-model reversals indicate that format preference is not uniform across model families.}
  \label{fig:format_sensitivity}
\end{figure}

\paragraph{Recoverability is above chance in a subset of settings and can dissociate from accuracy.}
Linear probes trained on \(I\) predict eventual correctness with AUROC above chance in a subset of model--benchmark cells. Strikingly, representational separability can co-occur with degraded behavioral performance: in our matrix, Qwen exhibits high probe AUROC on ARC-Challenge while CoT accuracy drops substantially in the same setting. This dissociation supports a core claim of the intention-collapse framing: an informative pre-collapse signal may exist without being reliably converted into a clean final decision, especially under discrete-choice response formats.

\paragraph{Statistical scope.}
Given the per-cell sample size (\(n=200\)), we primarily emphasize qualitative cross-regime patterns and large effect sizes. Where confidence intervals are reported (e.g., for probe AUROC), they are computed via bootstrap. More extensive paired hypothesis testing for accuracy differences (e.g., McNemar on per-item outcomes) is left to future work.

\begin{figure}[t]
  \centering
  \includegraphics[width=0.92\linewidth]{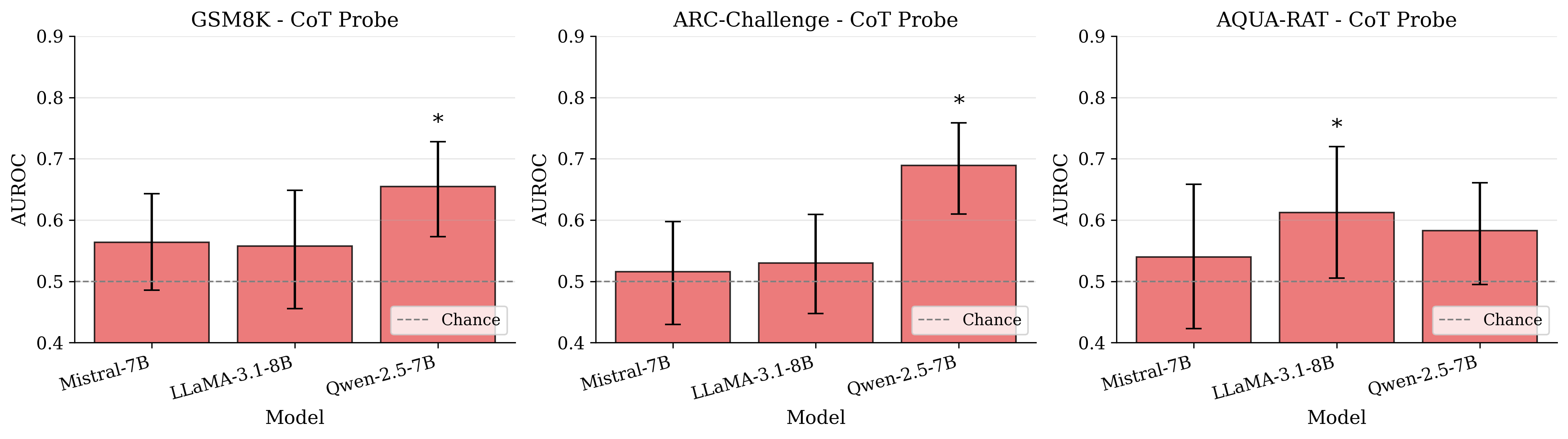}
  \caption{Probe AUROC (with 95\% bootstrap confidence intervals) predicting correctness from the pre-collapse intention state \(I\). Asterisks indicate CIs excluding 0.5.}
  \label{fig:probe_auroc}
\end{figure}

\begin{figure}[t]
  \centering
  \includegraphics[width=0.92\linewidth]{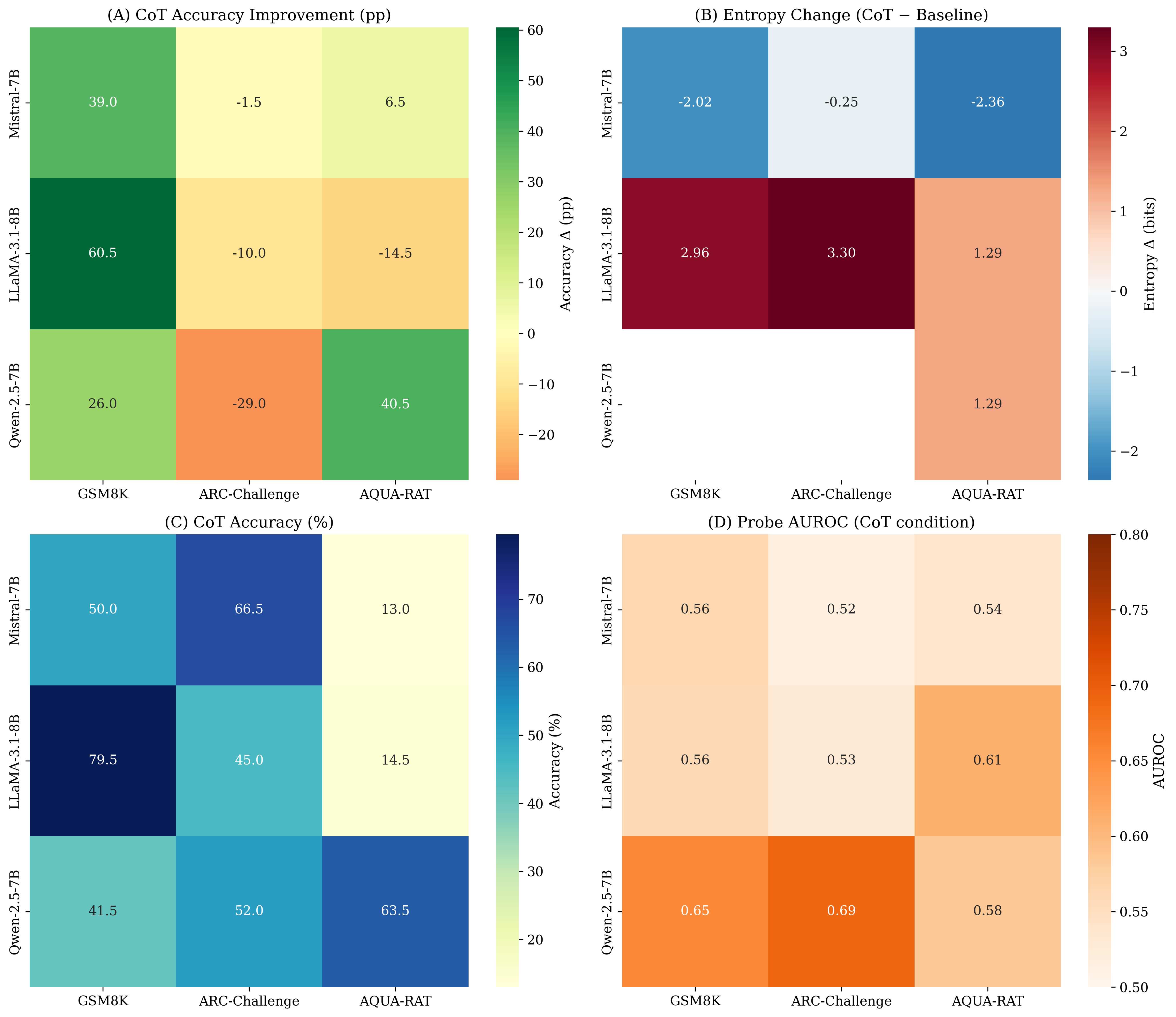}
  \caption{Heatmaps summarizing the \(3\times3\) matrix: (A) CoT accuracy improvement, (B) entropy regime shift \(\Delta H\), (C) CoT accuracy, and (D) probe AUROC.}
  \label{fig:heatmaps}
\end{figure}

\paragraph{Babble control: length alone does not explain the signatures.}
Babble produces long outputs but does not reliably reproduce the same metric--performance relationships observed under CoT. This supports the interpretation that CoT changes more than verbosity: it reshapes the pre-collapse state and/or alters the reliability of final commitment, in ways that are both model- and format-dependent.

\paragraph{Compute confound and multiple-choice compliance.}
CoT increases output length substantially relative to baseline, introducing a compute confound (additional decoding computation). Additionally, multiple-choice settings introduce a compliance axis (e.g., selecting a single valid option). We therefore treat response-format compliance as a necessary diagnostic for interpreting ARC/AQUA degradations: such drops may reflect genuine ``overthinking'' failures, extraction ambiguity, or instruction-following mismatches. We report compliance rates alongside accuracy in the final results tables.

\paragraph{Format-induced reasoning drift (``overthinking'') as commitment failure.}
The consistent ARC-Challenge degradations under CoT across all three model families suggest that externalizing intermediate reasoning is not uniformly beneficial under discrete-choice interfaces. One interpretation compatible with intention collapse is \emph{format-induced reasoning drift}: the model may enter a richer or more separable pre-collapse state, yet the subsequent collapse into a single option becomes less reliable when forced through an explicit verbal trajectory. This is consistent with cases where recoverability from $I$ remains above chance while CoT accuracy declines, indicating that latent task-relevant signal can exist without being faithfully expressed as a valid discrete commitment.

\subsection{Extensions: State-Dependent Collapse and Lossiness Diagnostics}
\label{sec:extensions}

The core study above establishes that intention metrics are measurable and reveal meaningful cross-model and cross-format structure. We now describe extensions that remain as forward-looking protocols for scaling the agenda.

\subsubsection{State-Dependent Collapse Variability}

Inspired by neuromodulation analogies (Section~\ref{sec:operationalizing}), we
test whether collapse variability adapts to internal state, as in RL-tuned
models like DeepSeek-R1 \parencite{deepseek2025r1}.

\textbf{Protocol.}
\begin{itemize}
    \item \textbf{Setup:} On GSM8K and MATH, run models with and without
    RL/TTT adaptations \parencite{deepseek2025r1, sun2020ttt}. For each instance,
    sample 10--20 outputs at varying temperatures (0.1 to 1.0), holding the
    intention phase fixed (e.g., via cached KV states).

    \item \textbf{Measurement:} Compute output variability as the average
    edit distance or semantic similarity (via embedding cosine) across
    samples. Condition on intention metrics: bin instances by high/low
    $H_{\text{int}}(I)$ or uncertainty estimates from layer norms. Fit a
    simple regressor (e.g., MLP) to predict ``optimal'' temperature from $I$
    features, then test if state-dependent decoding outperforms
    fixed-temperature baselines.

    \item \textbf{Analysis:} Compare variability distributions pre- and
    post-RL, and evaluate performance lifts from adaptive policies on
    held-out sets.

    \item \textbf{Hypotheses:} RL methods like DeepSeek-R1 should increase
    variability selectively in high-entropy states (encouraging exploration
    under uncertainty), mirroring biological adaptive gain
    \parencite{astonjones2005}. This could yield $3$--$7\%$ accuracy gains, as
    hinted in preliminary reports \parencite{wang2025mcts}, and validate
    state-dependent $\xi$ (Section~\ref{sec:operationalizing}).
\end{itemize}

This experiment bridges human-inspired modulation with LLM training, and is
testable in $\sim$5--10 GPU hours.

\subsubsection{Latent Knowledge Recovery Before vs.\ After Collapse}

To quantify collapse's lossiness, we leverage ``quirky'' models that
systematically err despite internal knowledge \parencite{burns2023quirky}.

\textbf{Protocol.}
\begin{itemize}
    \item \textbf{Setup:} Use or finetune quirky variants on GSM8K/ARC
    subsets where models know facts but output incorrectly (e.g., via LoRA
    to induce biases \parencite{burns2023quirky}).

    \item \textbf{Measurement:} For each instance, probe pre-collapse $I$
    (hidden states) vs.\ post-collapse outputs. Train probes on $I$ to
    recover ``true'' knowledge (e.g., correct answer, key lemma). Compare
    recoverability rates and mutual information between probed knowledge and
    verbalized content.

    \item \textbf{Analysis:} Stratify by error types (e.g., hallucination
    vs.\ reasoning slip) and conditions (baseline vs.\ refinement methods).

    \item \textbf{Hypotheses:} Pre-collapse recoverability should exceed
    post-collapse by $20$--$50\%$ on quirky instances, with refinement
    (e.g., self-refine) narrowing the gap by enriching $I$ to make latent
    knowledge more expressible. This aligns with ELK goals
    \parencite{christiano2021elk} and recent benchmarks questioning pure output
    evaluation.
\end{itemize}

This setup requires $\sim$15 GPU hours for finetuning/probing and directly
tests the framework's core claim of lossy projection.

\subsubsection{Summary and Broader Agenda}

These protocols form a lean research agenda: start with off-the-shelf models,
scale to finetuned variants, and iterate based on observed signatures. Early
evidence from related works (e.g., higher-dimensional representations in
refined models \parencite{gonen2024dim}) suggests positive results, but full
implementation will clarify intention collapse's utility for diagnosing and
improving LLMs. Future extensions could include multi-modal settings or human
studies, positioning this as a bridge between cognitive science and AI
alignment.

\section{Related Work}
Our framework connects several active threads: latent knowledge and interpretability, pre-generation uncertainty and adequacy estimation, reasoning-enhancing inference methods, and inference-time compute scaling. We discuss each in turn, emphasizing how intention collapse provides a shared object of analysis---the pre-collapse state \(I\)---and how our empirical \(3\times3\) matrix (three models; three benchmarks spanning free-response and multiple-choice formats) sharpens the distinctions among these lines.

\subsection{Latent Knowledge and Interpretability}
Prior work has repeatedly demonstrated that language models encode more information in their internal activations than is directly expressed in their outputs. Mechanistic interpretability studies formalize transformer representations as high-dimensional feature spaces with sparse ``circuits'' implementing specific behaviors \parencite{elhage2021,conmy2023}. Work on eliciting latent knowledge (ELK) argues that truth-like features can be recoverable from hidden states even when the model's overt answers are wrong \parencite{christiano2021elk,burns2023quirky}.

Particularly relevant is Contrast-Consistent Search (CCS; \parencite{burns2022discovering}), which trains unsupervised linear probes to identify directions in activation space corresponding to truth vs.\ falsehood without labeled data. CCS highlights a core phenomenon aligned with our framing: models can ``know'' without reliably ``saying.'' Our \(\text{Recov}(I; Z)\) metric is conceptually similar but differs in two respects: (i) we use supervised probes targeting task-specific outcomes (e.g., correctness) rather than unsupervised truth directions, and (ii) we focus on a strict pre-collapse extraction point rather than arbitrary intermediate activations. Empirically, our results reinforce the interpretability--elicitation picture in a format-sensitive setting: we observe cases where probe AUROC is reliably above chance even when behavioral accuracy drops under multiple-choice formats, suggesting that internal separability and final discrete commitment can dissociate.

More recent work investigates how internal knowledge supports complex reasoning \parencite{xia2025lact} and whether chain-of-thought is faithful to the model's internal computations \parencite{lanham2023measuring,turpin2023language}. These studies document gaps between internal computation and externalized rationale---precisely the gap intention collapse formalizes, while directing attention to the pre-first-token state as a controlled comparison point.

\subsection{Pre-Generation Uncertainty and Adequacy Estimation}
A growing literature aims to predict, \emph{before} or \emph{during} generation, whether a model will succeed at a given task. This directly relates to our \(H_{\text{int}}(I)\) and \(\text{Recov}(I; Z)\), which are computed (or trained) on a pre-collapse state.

Semantic entropy \parencite{kuhn2023semantic} estimates uncertainty by sampling multiple outputs and clustering them by semantic equivalence; high entropy over clusters indicates unreliable generations. Unlike our \(H_{\text{int}}(I)\), which measures single-pass entropy at a specific pre-collapse moment, semantic entropy requires multiple forward passes and post-hoc clustering. The approaches are complementary: \(H_{\text{int}}(I)\) is cheap and isolates a local commitment distribution, while semantic entropy captures output-level diversity at higher compute cost. In our \(3\times3\) results, \(H_{\text{int}}(I)\) primarily behaves as a regime- and model-family signature (distinct CoT-induced entropy regimes across models) rather than a uniformly strong item-level predictor, aligning with the intuition that token-local uncertainty is not sufficient to characterize end-to-end success.

CLOTHO \parencite{stengel2024clotho} models input-side adequacy using density estimation over hidden states (LIHS), achieving AUROC \(\approx 0.7\) for predicting pass/fail before generation begins. CLOTHO's pre-generation focus aligns closely with our extraction protocol; the key difference is framing and scope. CLOTHO treats adequacy prediction as an end in itself, while intention collapse unifies adequacy-style diagnostics with an explicit collapse operator and with interventions that reshape the pre-collapse state. Our probe-based \(\text{Recov}(I; Z)\) plays a similar diagnostic role and can be benchmarked against density-based adequacy estimators.

\(P(\mathrm{IK})\) and calibration methods \parencite{kadavath2022language} train models to predict whether they can correctly answer a question, while work on self-evaluation \parencite{lin2022teaching} and verbalized confidence \parencite{tian2023just} explores whether models can assess their own uncertainty. These output-side methods contrast with our focus on pre-collapse representations and with cases where internal signal is present but not reliably expressed in the final answer.

Early-exit policies in long-form reasoning \parencite{schuster2022confident,delcorro2023} use probes on intermediate hidden states to decide when to stop generation early, trading compute for accuracy. This work demonstrates that hidden states can carry linearly recoverable signals about eventual correctness---supporting the premise underlying \(\text{Recov}(I; Z)\)---and suggests practical applications of intention-level metrics for adaptive inference.

\subsection{Reasoning-Enhancing Inference-Time Methods}
A large body of work improves reasoning by modifying inference-time computation. Chain-of-thought prompting \parencite{wei2022chain} elicits step-by-step reasoning via few-shot examples or instructions. STaR \parencite{zelikman2022star} bootstraps reasoning by fine-tuning on self-generated rationales that lead to correct answers. Quiet-STaR \parencite{zelikman2024quietstar} extends this to internal thoughts that improve predictions without being shown to users. Process supervision and process reward models \parencite{lightman2023letsverify,uesato2022} provide step-level feedback. Self-refinement methods \parencite{madaan2023selfrefine} prompt critique and revision. Test-time training (TTT) \parencite{sun2020ttt,gandelsman2022tttmae} adapts parameters on the fly.

These methods are often studied in isolation, each with its own evaluation protocol. Our framework unifies them by treating each as an intervention on the intention state \(I\) prior to collapse, but our empirical findings caution against equating ``more reasoning'' with ``better answers.'' In particular, in our \(3\times3\) matrix, CoT improves accuracy strongly in free-response math while consistently degrading accuracy in multiple-choice abstract reasoning. This underscores that the benefit of a reasoning intervention depends on how it interacts with the response format and with the model's commitment policy---precisely the locus captured by the intention-collapse view.

\begin{table}[t]
\centering
\begin{tabular}{p{0.28\linewidth} p{0.66\linewidth}}
\hline
\textbf{Method} & \textbf{Primary effect on \(I\)} \\
\hline
Chain-of-thought & Conditions \(I\) toward structured intermediate states; may alter entropy regime and commitment reliability \\
STaR / Quiet-STaR & Trains richer intention trajectories; increases accessibility of latent variables relevant for prediction \\
Process reward models & Shapes \(I\) by rewarding valid intermediate paths through intention space \\
Self-refinement & Traverses multiple candidate intention states before final commitment \\
Test-time training & Adapts \(\theta\), making \(I(x;\theta)\) instance-specific \\
\hline
\end{tabular}
\caption{Reasoning-enhancing methods as interventions on the intention state prior to collapse.}
\label{tab:methods_interventions}
\end{table}

The role of our metrics is therefore not to assert that enriching \(I\) monotonically improves performance, but to enable direct comparison across interventions and to test whether changes in uncertainty, geometry, and recoverability mediate downstream outcomes across tasks and formats.

\subsection{Inference-Time Compute and Scaling}
Recent work on test-time compute scaling argues that allocating more computation at inference can rival or outperform scaling model parameters \parencite{snell2025scalingttc,brown2024}. This has prompted interest in how to allocate compute across search, verification, and refinement.

We build on this observation by separating \emph{compute that shapes \(I\)} from compute that only searches over post-collapse candidates. In our empirical matrix, CoT increases test-time computation via longer decoding, but the resulting accuracy gains are format-dependent and can reverse in multiple-choice settings. This motivates a more fine-grained hypothesis: the value of test-time compute may depend on whether it increases recoverability and yields a commitment policy that faithfully expresses that recoverable signal.

\subsection{Positioning of Our Contribution}
Table~\ref{tab:related} summarizes how our framework relates to prior work:
\begin{table}[t]
\centering
\begin{tabular}{p{0.30\linewidth} p{0.32\linewidth} p{0.32\linewidth}}
\hline
\textbf{Approach} & \textbf{Focus} & \textbf{Relation to intention collapse} \\
\hline
CCS \parencite{burns2022discovering} & Unsupervised truth probes & Supports lossy collapse; we use supervised probes at a strict pre-collapse point \\
CLOTHO \parencite{stengel2024clotho} & Pre-generation adequacy & Similar pre-generation goal; we embed it in a collapse-centric framework and compare across regimes \\
Semantic entropy \parencite{kuhn2023semantic} & Output-level uncertainty & Complementary; we measure single-pass pre-collapse entropy \\
Early-exit probes \parencite{schuster2022confident} & Adaptive inference & Demonstrates recoverability; we operationalize it as \(\text{Recov}(I; Z)\) at the first-step boundary \\
CoT / STaR / PRM & Reasoning enhancement & We unify as interventions on \(I\), while highlighting format-dependent failures of commitment \\
Inference scaling & Compute allocation & We motivate separating compute that reshapes \(I\) from post-collapse search \\
\hline
\end{tabular}
\caption{Positioning of intention collapse relative to related research threads.}
\label{tab:related}
\end{table}

Our contribution is not to replace these methods, but to provide a unifying lens---intention collapse---that connects them conceptually and offers simple, comparable metrics for characterizing how interventions reshape internal states and how those states do (or do not) translate into reliable discrete commitments across response formats.

\section{Limitations and Scope}
\label{sec:limitations}
We present intention collapse as a conceptual and empirical framework for studying the boundary between latent computation and discrete linguistic commitment, not as a complete theory of LLM reasoning. Several limitations bound the interpretation of our results and define concrete directions for strengthening the empirical case.

\subsection{Scope of Empirical Validation}
\paragraph{Model coverage.}
Our experiments span three instruction-tuned model families in the 7--8B parameter range (Mistral-7B-Instruct, Llama-3.1-8B-Instruct, Qwen-2.5-7B-Instruct), evaluated under a fixed greedy decoding policy. This scope enables controlled cross-model comparison, but it does not address scaling effects (e.g., 70B+), base (non-instruct) checkpoints, or substantially different architectures (e.g., mixture-of-experts, state-space models). We also evaluate in 4-bit quantization for practicality; quantization can alter activation statistics and may shift absolute metric values even when qualitative trends persist.

\paragraph{Task and format coverage.}
Our reported matrix uses three benchmarks: GSM8K (free-response math), ARC-Challenge (multiple-choice abstract reasoning), and AQUA-RAT (multiple-choice math). This design intentionally mixes domains and response formats and reveals strong format dependence: CoT can improve free-response accuracy while degrading multiple-choice accuracy. However, these benchmarks do not cover several important capability classes (e.g., factual retrieval, code synthesis, open-ended generation, multi-turn dialogue), where the mapping from \(I\) to output may behave differently. Moreover, differences between datasets (difficulty, distributional quirks, annotation conventions) can confound interpretations of ``format sensitivity'' unless paired-format controls are introduced.

\paragraph{Sample size and statistical power.}
We evaluate \(n=200\) items per model--benchmark cell, which is sufficient to detect some robust effects (e.g., large accuracy shifts; AUROC above chance in several settings) but limits fine-grained conclusions about small differences, layer-wise variability, or interaction effects among model family, benchmark, and regime. We therefore emphasize qualitative cross-condition patterns and report uncertainty via bootstrap confidence intervals.

\paragraph{Modality.}
Although the framework is modality-agnostic in principle (Section~2), all experiments are text-only. Extending the extraction protocol and metrics to vision-language models, document understanding, or multimodal instruction following remains future work.

\subsection{Limitations of the Proposed Metrics}
\paragraph{Intention entropy \(H_{\text{int}}(I)\).}
Across the \(3\times3\) matrix, \(H_{\text{int}}(I)\) behaves most reliably as a regime- and model-family signature (e.g., CoT induces distinct entropy \emph{regimes} across models when compared to baseline). However, item-level predictive power is weak and not monotonic with accuracy: cases exist where \(\Delta H\) changes substantially without corresponding gains in correctness.

Crucially, \(H_{\text{int}}(I)\) is an \emph{unconstrained} next-token entropy over the full vocabulary at the pre-first-token boundary. As such, it is sensitive to prompt phrasing and to lexical/tokenization details near the end of the prompt (including boilerplate CoT preambles), which can shift the next-token distribution for reasons unrelated to problem difficulty. Accordingly, we scope our interpretation of \(\Delta H\) to changes in \emph{pre-collapse generative dispersion} rather than calibrated decision uncertainty.

For multiple-choice benchmarks, more decision-aligned alternatives would compute entropy restricted to option tokens (e.g., \(\{A,B,C,D,E\}\)) and report pre-collapse logit margins between options. These require logging per-option log-probabilities at the same extraction boundary. Our current checkpoint format stores aggregated entropy values but not per-option log-probabilities, so we do not report option-normalized entropy or option margins in this revision; we instead treat this as a concrete logging improvement for future releases.

For MCQ tasks, decision-space-aligned diagnostics (option-normalized entropy over \(\{A,B,C,D,E\}\) and pre-collapse option margins) require per-option log-probabilities at the extraction boundary. Our current checkpoint format stores aggregated entropy values but not per-option log-probabilities; we therefore do not report option-normalized entropy or margins in this revision. This logging upgrade is a priority for future releases.

\paragraph{Effective dimensionality \(d_{\text{eff}}(I)\).}
The participation ratio summarizes spectral concentration but is insensitive to other geometric properties (e.g., clustering, curvature, anisotropy). Estimates can also be unstable at small \(N\), and in our setting \(d \gg N\) implies a finite-sample rank ceiling \(\mathrm{rank}(C^{(\ell)}) \le N-1\) that can bias spectrum-based estimates. We mitigate this by computing \(d_{\text{eff}}^{(\ell)}\) per layer and aggregating across layers (rather than concatenating layers for global PCA), and by reporting subsampling stability across multiple \(N\) values. Future work should pair participation ratios with complementary geometry probes (e.g., local intrinsic dimensionality estimators, clustering indices, or representational similarity analyses) and evaluate sensitivity to quantization and normalization choices.

\paragraph{Recoverability \(\text{Recov}(I; Z)\).}
Linear probes are deliberately conservative: they provide interpretability and reduce overfitting risk, but may underestimate information present in \(I\). Probe performance also varies by layer, and selecting the best layer introduces researcher degrees of freedom. We mitigate this with validation-only selection and by reporting fixed layer sets; nevertheless, probe results should be interpreted as evidence of \emph{linear accessibility} rather than an upper bound on latent information. To further guard against spurious recoverability in the high feature-to-sample regime, we additionally report train--test gaps and label-shuffle baselines (Appendix~\ref{app:probe_robustness}).

A further nuance is highlighted by our results: high probe AUROC can co-occur with degraded behavioral accuracy in multiple-choice settings. This dissociation is informative for the intention-collapse framing, but it also complicates naive interpretations of \(\text{Recov}(I; Z)\) as a proxy for ``competence.'' Recoverability measures separability in representations; it does not guarantee that the model's collapse policy will express that signal as a valid final decision under a given response format. We therefore also report cross-regime transfer probes (train in one regime, test in another) as a diagnostic of representational shift (Appendix~\ref{app:probe_transfer}).

\subsection{Format, Parsing, and Compliance as Confounds}
Multiple-choice benchmarks introduce an additional failure axis beyond domain reasoning: the model must comply with a constrained output interface (select exactly one valid option). Observed degradations under CoT on ARC-Challenge and mixed outcomes on AQUA-RAT may reflect a combination of (i) genuine reasoning interference, (ii) instruction-following mismatch, (iii) parsing ambiguity, or (iv) distribution shift between datasets.

We treat extraction and compliance as first-class diagnostics: in the final results tables we report answer compliance rates (percentage of outputs containing a single valid option token) alongside accuracy. More robust parsing (e.g., constrained decoding over option tokens) would reduce this confound but would also modify the collapse operator, changing the object of study. We view this as an important design trade-off for future work: intervention on collapse policy vs.\ measurement under unconstrained generation.

\paragraph{Decision-aligned MCQ diagnostics (logging scope).}
For multiple-choice tasks, decision-space-aligned diagnostics such as option-normalized entropy over the option set (e.g., \(\{A,B,C,D,E\}\)) and pre-collapse option margins (top-1 vs.\ top-2 logit gaps restricted to options) require per-option log-probabilities at the extraction boundary. Our current checkpoint format stores aggregated full-vocabulary entropy and pre-collapse hidden states, but not per-option log-probabilities; we therefore do not report option-normalized entropy or option margins in this revision. We instead emphasize compliance diagnostics and cross-regime representation analyses, and treat MCQ decision-aligned logging as a priority upgrade for future releases.

\subsection{Compute Confound and the Nature of CoT Gains}
CoT typically increases output length relative to baseline, introducing a compute confound: part of the gain may arise from additional decoding computation rather than solely from changes in the pre-collapse state \(I\). Our Babble condition is a partial control for verbosity and compliance under extended generation, but it does not equalize the \emph{semantic} structure of generated tokens, nor does it causally control pre-collapse metrics via output length since \(I\) is extracted before the first token is emitted. Stronger controls include length-matched structured reasoning, compute-matched self-consistency, or constrained-token budgets that equalize decoding cost across regimes. Disentangling ``more compute'' from ``different pre-collapse regime'' remains a central methodological challenge for interpreting CoT effects.

\paragraph{Self-consistency and decoding as part of the collapse operator.}
Our experiments fix greedy decoding to isolate prompt-induced shifts at a controlled collapse policy. Self-consistency replaces a single collapse with multiple sampled collapses followed by external aggregation (e.g., majority vote), effectively modifying the collapse operator. In the intention-collapse framing, this is not merely a stronger baseline but a different collapse policy; evaluating intention-level metrics under sampling-and-aggregation policies is a natural extension, but is outside the scope of this checkpointed greedy-decoding study.

\subsection{Correlation vs.\ Causation}
Our findings are correlational. We observe that different prompting regimes are associated with systematic shifts in \(H_{\text{int}}(I)\), \(d_{\text{eff}}(I)\), and \(\text{Recov}(I; Z)\), and that these shifts interact with response format and model family. However, we do not establish that changes in these metrics \emph{cause} accuracy gains or failures. Causal claims would require interventions on \(I\) (e.g., activation steering, representation editing, or controlled perturbations) and measurement of downstream changes in collapse outcomes.

\subsection{Extraction Point and the Definition of Pre-Collapse}
We define pre-collapse as the latent state at the first decoding step, before the first output token is selected. This operationalization is precise and reproducible, but it is not the only reasonable boundary. Alternative definitions could aggregate information across prompt processing, analyze attention-head dynamics, or model a short preamble window rather than a single position. Our choice prioritizes comparability across regimes, but it may miss dynamics that unfold after the first token or during extended generation. Future work should explore multi-point extraction protocols while preserving clear separation between pre- and post-collapse analyses.

\subsection{Comparisons to Existing Diagnostics}
We implement and report standard behavioral baselines (e.g., generated-token length and compliance) and show that intention-level metrics provide complementary signal in some settings. However, we do not exhaustively benchmark against all related diagnostics, including semantic entropy with clustering, CLOTHO-style LIHS density estimation, CCS-style unsupervised truth probes, or self-consistency and verifier-based pipelines. Such comparisons would strengthen conclusions about when intention metrics add unique value versus when they replicate existing adequacy and uncertainty measures.

\paragraph{Implemented baselines.}
Unconstrained next-token entropy \(H_{\text{int}}\), generated-token length, and output compliance (MCQ). We do not report option-token margins in this revision because per-option log-probabilities at the extraction boundary were not stored in the current checkpoints.

\paragraph{Planned comparisons.}
Semantic entropy, CLOTHO-style LIHS density, CCS unsupervised probes, self-consistency/verifiers, and option-normalized entropy/margins for MCQ under improved logging.

\section{Discussion and Future Directions}
Despite the limitations in Section~6, our results suggest that intention collapse is a useful organizing concept for studying the boundary between latent computation and discrete linguistic commitment. The framework's value lies less in any single metric than in the shift it enables: treating the pre-collapse intention state \(I\) as a first-class empirical object that can be compared across inference regimes, model families, and response formats.

\paragraph{What the \(3\times3\) matrix adds.}
Across three model families (Mistral-7B, Llama-3.1-8B, Qwen-2.5-7B) and three benchmarks spanning free-response and multiple-choice formats (GSM8K, ARC-Challenge, AQUA-RAT), we observe three stable themes.

First, chain-of-thought prompting is \emph{not} a universal improvement: it yields strong gains in free-response math while consistently degrading performance in multiple-choice abstract reasoning, and it produces mixed outcomes in multiple-choice math. This supports a format-dependent view of reasoning interventions: success depends not only on domain competence but on whether the collapse policy reliably converts latent computation into a valid discrete decision under the required output interface.

Second, CoT induces distinct \emph{entropy regimes} across model families relative to baseline. Some models operate at lower pre-collapse entropy under CoT, while others operate at higher pre-collapse entropy. These regimes reflect different degrees of token-level uncertainty at the commitment boundary, and they are not uniformly predictive of correctness. This heterogeneity cautions against one-size-fits-all narratives of what CoT ``does'' internally.

Third, we observe a principled dissociation between internal signal and behavioral success: probe AUROC for predicting correctness from \(I\) can be reliably above chance even when accuracy declines under CoT in multiple-choice settings. In the language of intention collapse, this corresponds to a failure mode where information is present and linearly accessible pre-collapse, yet the mapping from \(I\) to the final committed output is unreliable under the response format. Put simply: \emph{knowing is not the same as committing}.

\paragraph{Toward actionable diagnostics: policy at the collapse boundary.}
A natural next step is to move from diagnostic measurement to interventions that directly target the collapse boundary.
\begin{itemize}
  \item \textbf{State-dependent decoding.} Learn or calibrate policies \(\pi(\xi \mid I)\) that adapt sampling behavior (or commitment rules) based on \(H_{\text{int}}(I)\) and probe confidence. In high-uncertainty intention states, exploration may be beneficial; in low-uncertainty states, conservative commitment may reduce avoidable errors.
  \item \textbf{Format-aware collapse policies.} For multiple-choice tasks, enforce constrained decoding over valid option tokens or apply structured post-processing that separates reasoning text from the final discrete choice. This directly addresses compliance/parsing confounds and tests whether accuracy drops arise from reasoning interference or from unreliable commitment under format constraints.
  \item \textbf{Early-exit and reroute.} Use \(\text{Recov}(I; Z)\) or calibration-aware probe scores to predict impending failure early and reroute computation to self-consistency, verification, or refinement only when needed, rather than uniformly increasing test-time compute.
\end{itemize}

\paragraph{Methodological strengthening.}
Several methodological improvements would sharpen claims and reduce confounds:
\begin{itemize}
  \item \textbf{Compute-matched controls.} Replace Babble with controls that match decoding cost while preserving task relevance (e.g., length-matched structured reasoning or compute-matched self-consistency), to better separate ``more compute'' from changes in pre-collapse state.
  \item \textbf{Compliance as a primary outcome.} Report option-validity and extraction success rates alongside accuracy for multiple-choice benchmarks, and analyze whether intention metrics predict compliance failures distinctly from reasoning failures.
  \item \textbf{Multi-point extraction.} Extend beyond a single pre-collapse boundary by extracting representations at standardized anchors (e.g., immediately after an ``Answer:'' delimiter) and by separating pre-collapse from early post-collapse dynamics to test whether information loss occurs abruptly or progressively.
\end{itemize}

\paragraph{Connections to interpretability and safety.}
The observation that models can exhibit recoverable internal signal that is not reliably expressed echoes a central theme in interpretability and alignment: models may ``know more than they say.'' Intention collapse provides a compact vocabulary for studying this gap at a well-defined boundary, complementing work on latent knowledge discovery \parencite{burns2022discovering} and analyses of chain-of-thought faithfulness \parencite{lanham2023measuring}. Beyond diagnostics, it suggests an actionable research direction: alignment and reliability may hinge not only on what is represented internally, but on how commitment policies translate representations into decisions under varied interfaces and constraints.

\paragraph{Closing perspective.}
Intention collapse is ultimately a proposal about structure: rich internal dynamics followed by an irreversible projection onto a thin linguistic surface. By making the pre-collapse state measurable and comparable, we aim to turn that structural analogy into a practical research program---one that explains why reasoning interventions help in some settings, fail in others, and sometimes reveal a deeper gap between latent competence and expressed behavior.

\section{Conclusion}
\label{sec:conclusion}

We proposed \emph{intention collapse} as a unifying lens for analyzing language
generation in both humans and large language models. Rather than treating an
utterance as an opaque token sequence, we separate a high-dimensional intention
state \(I\)—formed after prompt conditioning and internal computation—from an
irreversible collapse operator \(\kappa_\theta\) that maps \(I\) into a discrete
linguistic trajectory. The human analogy is deliberately coarse: we do not claim
that LLMs have conscious intentions, but we argue that both biological and
artificial systems share a structural pattern of rich internal dynamics followed
by a sharply reduced verbal surface.

Within this framework, we reinterpret a range of contemporary reasoning
techniques—chain-of-thought prompting, STaR, Quiet-STaR, process reward models,
test-time training, self-refinement, and search-based decoding—as interventions
on \(I\) enacted before collapse. This perspective enables comparisons among
methods often studied in isolation and shifts the central question from ``does it
work?'' to ``what changes at the commitment boundary?''

To operationalize the framework, we introduced three simple, model-agnostic
metrics at the pre-collapse boundary: intention entropy \(H_{\text{int}}(I)\),
effective dimensionality \(\dim_{\text{eff}}(I)\), and latent recoverability
\(\mathrm{Recov}(I; Z)\). These are computable with standard tools (entropy of
the predictive distribution, PCA on activations, and linear probes) and provide
a compact vocabulary for describing how much task-relevant structure is present
before the first token is emitted.

We instantiated the framework in a controlled \(3\times3\) empirical study
spanning three instruction-tuned model families (Mistral-7B, Llama-3.1-8B,
Qwen-2.5-7B), three benchmarks spanning free-response and multiple-choice
formats (GSM8K, ARC-Challenge, AQUA-RAT), and three inference regimes (baseline,
CoT, and a verbosity-matched babble control), using \(n=200\) items per
model--benchmark cell. Three findings stand out. First, the impact of CoT is
strongly format-dependent: CoT yields large gains in free-response math yet can
degrade performance in multiple-choice settings, underscoring that additional
reasoning tokens do not guarantee a reliable discrete commitment under a
constrained output interface. Second, CoT induces distinct entropy regimes
across model families relative to baseline—some models operate at lower
pre-collapse entropy under CoT while others operate at higher entropy—showing
that CoT does not have a single universal internal signature. Third,
recoverability from \(I\) can dissociate from behavioral success: probe AUROC
for predicting correctness from the pre-collapse state can be above chance even
when accuracy declines, suggesting that informative internal signal may exist
without being consistently expressed in the final answer. Taken together, these
results support the core intuition of intention collapse: performance is shaped
not only by what is represented in \(I\), but by how reliably collapse
transforms that representation into a valid decision.

\paragraph{Engineering relevance.}
Beyond diagnosis, intention-level metrics suggest a pragmatic control surface for
test-time compute. Because \(H_{\text{int}}(I)\), \(\dim_{\text{eff}}(I)\), and
\(\mathrm{Recov}(I; Z)\) are computed at the pre-collapse boundary, they can be
used to route inference before expensive generation: for example, to decide when
to use direct answering versus CoT, when to invoke verification or
self-consistency, and—critically for multiple-choice formats—when to enforce
format-aware commitment (e.g., constrained decoding over valid options). This
framing turns intention collapse into a concrete proposal for adaptive inference:
allocate additional computation only when the pre-collapse state indicates
uncertainty or a risk of commitment failure.

At the same time, our contribution is intentionally modest in scope. We do not
offer a psychological theory of human intention or a mechanistic account of all
internal computations in transformers. Instead, we provide a mesoscale tool that
is precise enough to be instantiated and falsified on contemporary models, yet
broad enough to connect interpretability, uncertainty estimation, and
inference-time compute. Future work should strengthen controls (especially
compute-matching and multiple-choice compliance), extend extraction beyond a
single boundary point, and test whether intention-level diagnostics can guide
actionable policies (e.g., state-dependent decoding and format-aware commitment
rules). If successful, intention collapse may help turn ``reasoning'' from a
loose label into a measurable research target: systems whose commitment boundary
can be characterized, compared, and eventually controlled.

\paragraph{Practical hooks for adaptive inference.}
Because $\Hint(I)$, $\dimeff(I)$, and $\Recov(I;Z)$ are available \emph{before} expensive generation, they naturally support (i) \emph{routing} (direct answer vs.\ CoT vs.\ verify/self-consistency), (ii) \emph{format-aware commitment} (e.g., constrained decoding over valid option tokens in multiple-choice), and (iii) \emph{early rerouting} when intention-state diagnostics indicate high uncertainty or low commitment reliability. Importantly, these interventions act on the collapse boundary rather than only post-hoc filtering of outputs.

\clearpage

\appendix

\section{Experimental Details}
\label{app:experimental-details}

\subsection{Code Availability}
All code for reproducing the experiments in this paper is available at:
\begin{center}
\url{https://github.com/patriciomvera/intention-collapse-experiments}
\end{center}

The repository includes: (i) scripts for activation extraction at the pre-collapse
boundary (Section~2), (ii) metric computation for \(H_{\text{int}}(I)\) and
\(\dim_{\text{eff}}(I)\), (iii) probe training and evaluation for
\(\mathrm{Recov}(I;Z)\), (iv) parsers and compliance diagnostics for multiple-choice
benchmarks, and (v) notebooks to reproduce figures and consolidated tables for the
\(3\times3\) matrix reported in Section~\ref{sec:experiments}.

We release code and prompt templates sufficient to reproduce all reported tables and figures at the accompanying repository.

\subsection{Models and Benchmarks}
We evaluate three instruction-tuned open-weight models in the 7--8B range:
\begin{itemize}
  \item Mistral-7B-Instruct
  \item Llama-3.1-8B-Instruct
  \item Qwen-2.5-7B-Instruct
\end{itemize}
All runs use greedy decoding (temperature 0.0) and practical inference settings
(4-bit quantization for activation extraction). For each model--benchmark cell we
evaluate \(n=200\) randomly sampled items.

Benchmarks:
\begin{itemize}
  \item \textbf{GSM8K (free-response math)}: numeric final answers.
  \item \textbf{ARC-Challenge (multiple-choice)}: one option among a fixed set.
  \item \textbf{AQUA-RAT (multiple-choice math)}: one option among a fixed set.
\end{itemize}

\subsection{Prompt Templates}
\label{app:prompts}

We used separate prompt templates for free-response (GSM8K) and multiple-choice
(ARC/AQUA) to ensure that the commitment format is explicit. All prompts are
applied within the same three regimes: Baseline, CoT, and Babble.

\paragraph{GSM8K: Baseline (Direct Answer).}
\begin{verbatim}
Solve this math problem. Give only the final
numerical answer.

Problem: {question}
Answer:
\end{verbatim}

\paragraph{GSM8K: CoT (Chain-of-Thought).}
\begin{verbatim}
Solve this math problem step by step. Show your
reasoning, then give the final answer after ####.

Problem: {question}
Solution:
\end{verbatim}

\paragraph{GSM8K: Babble (Length Control).}
\begin{verbatim}
Given this math problem, write a long stream of
consciousness about numbers, calculations, and
mathematical concepts. Do NOT solve the problem -
just write loosely related mathematical musings
for about 100 words.

Problem: {question}
Stream of consciousness:
\end{verbatim}

\paragraph{Multiple-choice: Baseline (Direct Answer).}
\begin{verbatim}
Choose the correct option. Reply with only one
letter from the available choices.

Question: {question}
Choices: {choices}
Answer:
\end{verbatim}

\paragraph{Multiple-choice: CoT (Chain-of-Thought).}
\begin{verbatim}
Think step by step, then choose the correct option.
At the end, reply with only one letter from the
available choices.

Question: {question}
Choices: {choices}
Reasoning:
Final answer:
\end{verbatim}

\paragraph{Multiple-choice: Babble (Length Control).}
\begin{verbatim}
Write a long stream of consciousness about the topic
and the possible answers. Do NOT solve the problem.
At the end, still output exactly one letter from the
available choices.

Question: {question}
Choices: {choices}
Stream of consciousness:
Final answer:
\end{verbatim}

\subsection{Answer Extraction, Parsing, and Compliance}
For GSM8K, we extract the final numeric answer using standard normalization rules
(e.g., last occurring numeric token in the final answer span after the delimiter).
For multiple-choice benchmarks, we extract the final option letter using a strict
parser (single valid option among the allowed set). If parsing fails (no valid
single option), the output is marked incorrect. To diagnose format effects, we
report \textbf{compliance rates} for multiple-choice tasks (percentage of generations
containing a single valid option token at the end) alongside accuracy. This is
particularly important in settings where CoT reduces accuracy, since some failures
may reflect commitment/compliance breakdown rather than lack of latent signal.

\subsection{Decoding Parameters}
\label{app:decoding}

All three conditions used greedy decoding to eliminate stochasticity:
\begin{itemize}
    \item \textbf{Temperature}: 0.0 (greedy decoding)
    \item \textbf{Max tokens (Baseline)}: 50
    \item \textbf{Max tokens (CoT/Babble)}: 512
\end{itemize}

Greedy decoding ensures that differences in \(H_{\text{int}}(I)\),
\(\dim_{\text{eff}}(I)\), and \(\mathrm{Recov}(I;Z)\) are attributable to prompt
conditioning and the resulting generation regime rather than sampling randomness.
At the same time, the larger budgets for CoT/Babble introduce a compute confound;
we therefore interpret length controls as partial, and we treat compute-matching
as a priority for future work (Section~6 and Section~7).

\subsection{Pre-collapse Extraction and Probe Protocol (Summary)}
We extract \(I\) at the last prompt position immediately before the first output
token is selected, identically across Baseline, CoT, and Babble. We compute:
(i) \(H_{\text{int}}(I)\) from the next-token distribution at that position;
(ii) \(\dim_{\text{eff}}(I)\) via per-layer PCA participation ratios with
variance-weighted aggregation; and (iii) \(\mathrm{Recov}(I;Z)\) via
\(\ell_2\)-regularized logistic regression probes with item-level splits
(60/20/20) and bootstrap confidence intervals on AUROC. Additional robustness
checks for \(d_{\text{eff}}\) (subsampling and intrinsic-dimension cross-checks)
and for probes (train--test gaps, label-shuffle baselines, and cross-regime
transfer) are reported in Appendix~\ref{app:robustness}.

\section{Robustness and Additional Analyses}
\label{app:robustness}

\subsection{Effective dimensionality robustness}
\label{app:deff_robustness}

\begin{figure}[t]
  \centering
  \includegraphics[width=0.95\linewidth]{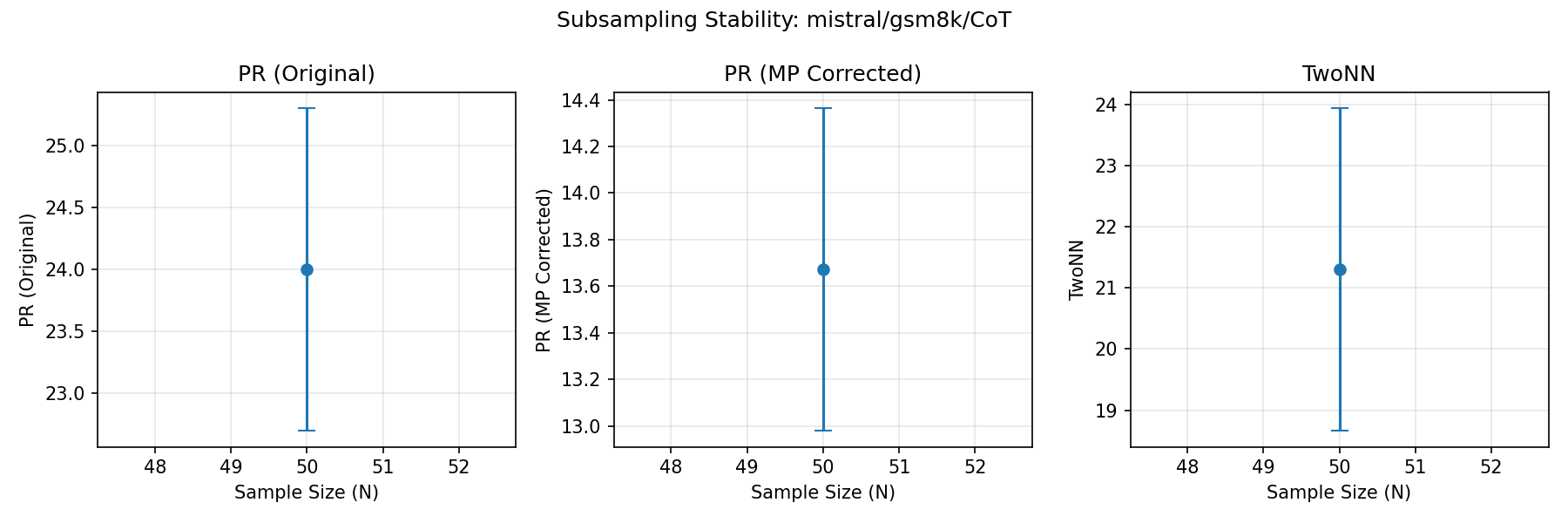}
  \caption{Subsampling stability of \(d_{\text{eff}}^{\text{agg}}\) across item counts \(N'\). Error bars indicate variation across repeated subsamples.}
  \label{fig:deff_subsampling}
\end{figure}

\paragraph{Subsampling stability.}
To assess finite-sample sensitivity in the \(d \gg N\) regime, we recompute \(d_{\text{eff}}\) under repeated subsampling of items within each model--benchmark--regime cell. Specifically, for each \(N' \in \{25, 50, 100, 150, 200\}\), we sample \(N'\) items without replacement (repeated \(R\) times) and report the mean and standard deviation of \(d_{\text{eff}}^{\text{agg}}\). Figure~\ref{fig:deff_subsampling} shows that the qualitative cross-regime trends emphasized in the main text are stable across subsample sizes.

\paragraph{Cross-checks with intrinsic-dimension estimators.}
Participation ratio is spectrum-based and can be biased by finite-sample covariance estimation. As complementary checks, we compare trends against intrinsic-dimension estimators derived from local neighborhood structure (TwoNN) and maximum-likelihood estimators (Levina--Bickel). While these estimators differ in absolute scale and assumptions, they corroborate the direction of regime shifts in the conditions where \(d_{\text{eff}}\) changes most strongly.

\subsection{Probe robustness checks}
\label{app:probe_robustness}

We evaluate probe robustness in three ways: (i) train--test AUROC gaps under the selected regularization strength \(C\), (ii) a label-shuffle sanity check (training labels permuted; expected AUROC \(\approx 0.5\)), and (iii) reporting a fixed pre-specified layer set to limit researcher degrees of freedom.

\begin{table}[h]
\centering
\caption{Probe robustness analysis. Train and test AUROC with 95\% bootstrap confidence intervals, label-shuffle sanity check (expected $\approx 0.5$), and train--test gap (positive values indicate potential overfitting).}
\label{tab:probe_robustness}
\small
\begin{tabular}{llccccc}
\toprule
Model & Benchmark & Condition & Train & Test [95\% CI] & Shuffle & Gap \\
\midrule
Mistral & GSM8K & baseline & 1.00 & 0.57 [0.37, 0.79] & 0.53 & +0.43 \\
Mistral & GSM8K & enhanced & 0.98 & 0.51 [0.35, 0.66] & 0.51 & +0.47 \\
Mistral & ARC & baseline & 1.00 & 0.86 [0.75, 0.95] & 0.53 & +0.14 \\
Mistral & ARC & enhanced & 1.00 & 0.48 [0.32, 0.63] & 0.54 & +0.52 \\
Mistral & AQUA & baseline & 1.00 & 0.74 [--, --] & 0.57 & +0.26 \\
Mistral & AQUA & enhanced & 1.00 & 0.75 [0.58, 0.89] & 0.57 & +0.25 \\
Llama & GSM8K & baseline & 1.00 & 0.73 [0.53, 0.91] & 0.49 & +0.27 \\
Llama & GSM8K & enhanced & 1.00 & 0.50 [0.32, 0.68] & 0.60 & +0.50 \\
Llama & ARC & baseline & 0.95 & 0.56 [0.42, 0.71] & 0.52 & +0.39 \\
Llama & ARC & enhanced & 1.00 & 0.50 [0.36, 0.66] & 0.49 & +0.50 \\
Llama & AQUA & baseline & 0.97 & 0.63 [0.46, 0.79] & 0.53 & +0.33 \\
Llama & AQUA & enhanced & 0.98 & 0.46 [0.30, 0.64] & 0.57 & +0.52 \\
Qwen & GSM8K & baseline & 1.00 & 0.75 [0.57, 0.88] & 0.55 & +0.25 \\
Qwen & GSM8K & enhanced & 0.97 & 0.60 [0.45, 0.75] & 0.49 & +0.37 \\
Qwen & ARC & baseline & 0.95 & 0.82 [0.71, 0.91] & 0.56 & +0.13 \\
Qwen & ARC & enhanced & 1.00 & 0.71 [0.56, 0.84] & 0.51 & +0.29 \\
Qwen & AQUA & baseline & 1.00 & 0.58 [0.36, 0.80] & 0.52 & +0.42 \\
Qwen & AQUA & enhanced & 0.95 & 0.62 [0.46, 0.75] & 0.52 & +0.34 \\
\bottomrule
\end{tabular}
\vspace{2pt}
{\footnotesize \emph{Note:} For severely imbalanced cells, naive bootstrap resamples may contain only one class, making AUROC undefined; we report CI as unavailable in those cases.}

\end{table}

\paragraph{Note on bootstrap CIs under severe imbalance.}
For highly imbalanced cells (e.g., AQUA baseline with very few positives), naive bootstrap resamples may contain only one class, making AUROC undefined and yielding NaN confidence intervals. In such cases we omit the CI or report it as unavailable; this does not affect the point estimate and is consistent with standard AUROC practice under extreme imbalance.

\subsection{Cross-regime probe transfer}
\label{app:probe_transfer}

\begin{figure}[t]
  \centering
  \includegraphics[width=0.98\linewidth]{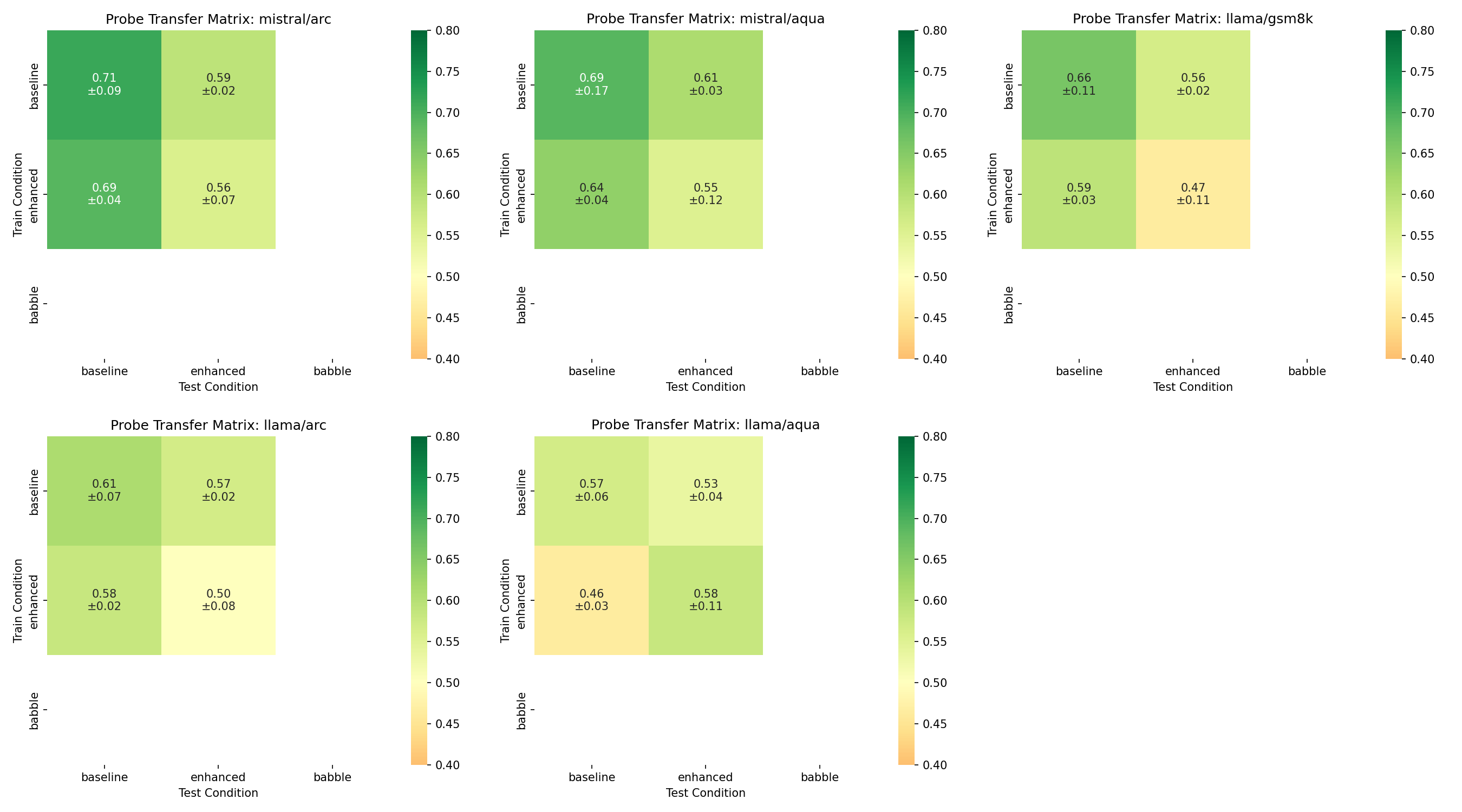}
  \caption{Cross-regime transfer matrices (AUROC). Systematic degradation under transfer indicates representational shift across regimes even when within-regime recoverability is above chance.}
  \label{fig:probe_transfer}
\end{figure}

To test whether recoverable information corresponds to shared predictive axes or regime-specific encodings, we evaluate cross-regime transfer: train on one regime and test on another while preserving item splits. Figure~\ref{fig:probe_transfer} summarizes transfer matrices (AUROC) for the available model/benchmark settings.

\subsection{Full per-cell numerical results}
\label{app:full_results}

For transparency, Table~\ref{tab:full_results} reports all per-cell values used to generate the main figures: accuracy, \(H_{\text{int}}\) mean/std, \(d_{\text{eff}}^{\text{agg}}\), generated-token counts, and probe AUROC with 95\% bootstrap confidence intervals.

\begin{table*}[t]
\centering
\caption{Main results across models and benchmarks. We report accuracy (\%), pre-collapse entropy $H_{\text{int}}(I)$ in bits, and probe AUROC with 95\% CI for the CoT condition. Arrows indicate change from Baseline to CoT.}
\label{tab:main_results}
\small
\begin{tabular}{llccccc}
\toprule
Model & Benchmark & Acc (B$\rightarrow$CoT) & $H_{\text{int}}$ (B$\rightarrow$CoT) & Pattern & AUROC [95\% CI] \\
\midrule
Mistral-7B & GSM8K & 11.0$\rightarrow$50.0 & 2.39$\rightarrow$0.37 & CF & 0.56 [0.49, 0.64] \\
 & ARC-Challenge & 68.0$\rightarrow$66.5 & 0.94$\rightarrow$0.69 & CF & 0.52 [0.43, 0.60] \\
 & AQUA-RAT & 6.5$\rightarrow$13.0 & 2.85$\rightarrow$0.49 & CF & 0.54 [0.42, 0.66] \\
\midrule
LLaMA-3.1-8B & GSM8K & 19.0$\rightarrow$79.5 & 1.57$\rightarrow$4.53 & EC & 0.56 [0.46, 0.65] \\
 & ARC-Challenge & 55.0$\rightarrow$45.0 & 0.98$\rightarrow$4.27 & EC & 0.53 [0.45, 0.61] \\
 & AQUA-RAT & 29.0$\rightarrow$14.5 & 3.27$\rightarrow$4.56 & EC & 0.61* [0.51, 0.72] \\
\midrule
Qwen-2.5-7B & GSM8K & 15.5$\rightarrow$41.5 & 0.67$\rightarrow$2.21 & EC & 0.65* [0.57, 0.73] \\
 & ARC-Challenge & 81.0$\rightarrow$52.0 & 0.67$\rightarrow$1.57 & EC & 0.69* [0.61, 0.76] \\
 & AQUA-RAT & 23.0$\rightarrow$63.5 & 1.24$\rightarrow$2.53 & EC & 0.58 [0.49, 0.66] \\
\bottomrule
\end{tabular}
\vspace{1mm}
\begin{flushleft}
\footnotesize{CF = Collapse-first (entropy decreases); EC = Explore-then-commit (entropy increases). * indicates AUROC CI excludes chance (0.5).}
\end{flushleft}
\end{table*}

\clearpage

\printbibliography

\end{document}